\pdfoutput=1
\documentclass[10pt,twocolumn,letterpaper]{article}

\usepackage{iccv}
\usepackage{times}
\usepackage{epsfig}
\usepackage{graphicx}
\usepackage{amsmath}
\usepackage{amssymb}

\usepackage{epigraph}
\DeclareMathSymbol{@}{\mathord}{letters}{"3B}

\edef\oldtt{\ttdefault}
\usepackage[scaled]{beramono}
\usepackage[T1]{fontenc}
\renewcommand{\ttdefault}{\oldtt}

\newcommand{\ourtitle}{Habitat: A Platform for Embodied AI Research\xspace}

\newcommand{\habitatsim}{\texttt{Habitat-Sim}\xspace}
\newcommand{\habitatapi}{\texttt{Habitat-API}\xspace}
\newcommand{\habitatchal}{Habitat Challenge\xspace}
\newcommand{\turnleft}{\texttt{turn\_left}\xspace}
\newcommand{\turnright}{\texttt{turn\_right}\xspace}
\newcommand{\forward}{\texttt{move\_forward}\xspace}
\newcommand{\callstop}{\texttt{stop}\xspace}
\newcommand{\maxsteps}{$500$\xspace}
\newcommand{\spl}{\texttt{SPL}\xspace}
\newcommand{\gdsp}{\texttt{GDSP}\xspace}
\newcommand{\blind}{\texttt{Blind}\xspace}
\newcommand{\rgb}{\texttt{RGB}\xspace}
\newcommand{\depth}{\texttt{Depth}\xspace}
\newcommand{\rgbd}{\texttt{RGBD}\xspace}

\newcommand{\xhdr}[1]{\vspace{2pt}\noindent\textbf{#1}}

\usepackage{paralist}
\usepackage{comment}
\usepackage{mathtools}
\usepackage{microtype}
\usepackage{algorithm}
\usepackage{algorithmic}
\usepackage{nicefrac}
\usepackage{subcaption}
\usepackage{mysymbols}
\usepackage{pythonhighlight}

\newlength{\sectionReduceTop}
\newlength{\sectionReduceBot}
\newlength{\subsectionReduceTop}
\newlength{\subsectionReduceBot}
\newlength{\abstractReduceTop}
\newlength{\abstractReduceBot}
\newlength{\captionReduceTop}
\newlength{\captionReduceBot}
\newlength{\subsubsectionReduceTop}
\newlength{\subsubsectionReduceBot}

\newlength{\eqnReduceTop}
\newlength{\eqnReduceBot}

\newlength{\horSkip}
\newlength{\verSkip}

\newlength{\figureHeight}
\usepackage{multirow}
\usepackage{rotating}
\usepackage{booktabs}
\usepackage{tabularx}

\usepackage{balance}

\usepackage[pagebackref=true,breaklinks=true,colorlinks,bookmarks=false]{hyperref}

\usepackage{cleveref}

\iccvfinalcopy

\begin{document}
\def\iccvPaperID{4367} 
\def\httilde{\mbox{\tt\raisebox{-.5ex}{\symbol{126}}}}

\ificcvfinal\pagestyle{empty}\fi

\author{
\textbf{Manolis Savva$^{1,4}$*, Abhishek Kadian$^{1}$*, Oleksandr Maksymets$^{1}$*, Yili Zhao$^{1}$,}\\[3pt]
\textbf{Erik Wijmans$^{1,2,3}$, 
Bhavana Jain$^{1}$, Julian Straub$^{2}$, Jia Liu$^{1}$, Vladlen Koltun$^{5}$,}\\[3pt]
\textbf{Jitendra Malik$^{1,6}$, Devi Parikh$^{1,3}$, Dhruv Batra$^{1,3}$}\\[12pt]
$^{1}$Facebook AI Research, $^{2}$Facebook Reality Labs, 
$^{3}$Georgia Institute of Technology,\\
$^{4}$Simon Fraser University, $^{5}$Intel Labs, $^{6}$UC Berkeley\\
\url{https://aihabitat.org}
\vspace{-20pt}
}
\title{\ourtitle}
\maketitle

\pagenumbering{gobble} 
\let\svthefootnote\thefootnote
\let\thefootnote\relax\footnote{\llap{\textsuperscript{*}}Denotes equal contribution.}
\addtocounter{footnote}{-1}\let\thefootnote\svthefootnote

\begin{abstract}
We present Habitat, a platform for research in embodied artificial intelligence (AI).
Habitat enables training embodied agents (virtual robots) in highly efficient photorealistic 3D simulation.
Specifically, Habitat consists of: \\
(i) \habitatsim: a flexible, high-performance 3D simulator with configurable agents, sensors, and generic 3D dataset handling. \habitatsim is fast -- when rendering a scene from Matterport3D, it achieves several thousand frames per second (fps) running single-threaded, and can reach over $10@000$ fps multi-process on a single GPU. \\ 
(ii) \habitatapi: a modular high-level library for end-to-end development of embodied AI algorithms -- defining tasks (\eg navigation, instruction following, question answering), configuring, training, and benchmarking embodied agents.

These large-scale engineering contributions enable us to answer scientific questions requiring experiments that were till now impracticable or `merely' impractical.
Specifically, in the context of point-goal navigation:
(1) we revisit the comparison between learning and SLAM approaches from two recent works~\cite{mishkin2019benchmarking, kojima2019learn} and find evidence for the \textbf{\emph{opposite conclusion}} --
that learning outperforms SLAM if scaled to an order of magnitude 
more experience than previous investigations, and
(2) we conduct the first cross-dataset generalization experiments 
$\{$train, test$\} \times \{$Matterport3D, Gibson$\}$ for multiple sensors $\{$blind, RGB, RGBD, D$\}$ and find that only agents with depth (D) sensors generalize across datasets.
We hope that our open-source platform and these findings will advance research in embodied AI.

\end{abstract}
\section{Introduction}

\epigraph{The embodiment hypothesis is the idea that intelligence emerges in the interaction of an agent with
an environment
and as a result of sensorimotor activity.}
{\textit{Smith and Gasser~\cite{smith_al05}}}

Imagine walking up to a home robot and asking \myquote{Hey -- can you go check if my laptop is on my desk? And if so, bring it to me.} In order to be successful, such a robot would need a range of skills -- visual perception (to recognize scenes and objects), language understanding (to translate questions and instructions into actions), and navigation in complex environments (to move and find things in a changing environment). 

While there has been significant progress in the vision and language communities thanks to recent advances in deep representations~\cite{he_cvpr16, devlin_arxiv18}, much of this progress has been on `internet AI' rather than \emph{embodied} AI. The focus of the former is pattern recognition in images, videos, and text on \emph{datasets} typically curated from the internet~\cite{Deng09imagenet, mscoco, antol_iccv15}. The focus of the latter is to enable action by an embodied agent (\eg a robot) in an \emph{environment}. This brings to the fore active perception, long-term planning, learning from interaction, and holding a dialog grounded in an environment.

A straightforward proposal is to train agents directly in the physical world -- exposing them to all its richness.
This is valuable and will continue to play an important role in the development of AI.
However, we also recognize that training robots in the real world is 
\begin{inparaitem}[]
    \item \emph{slow} (the real world runs no faster than real time and cannot be parallelized), 
    \item \emph{dangerous} (poorly-trained agents can unwittingly injure themselves, the environment, or others), 
    \item \emph{resource intensive} (the robot(s) and the environment(s) in which they execute demand resources and time), 
    \item \emph{difficult to control} (it is hard to test corner-case scenarios as these are, by definition, infrequent and challenging to recreate), and 
    \item \emph{not easily reproducible} (replicating conditions across experiments and institutions is difficult). 
\end{inparaitem}

\begin{figure*}[t]
    \centering
    \includegraphics[width=\textwidth]{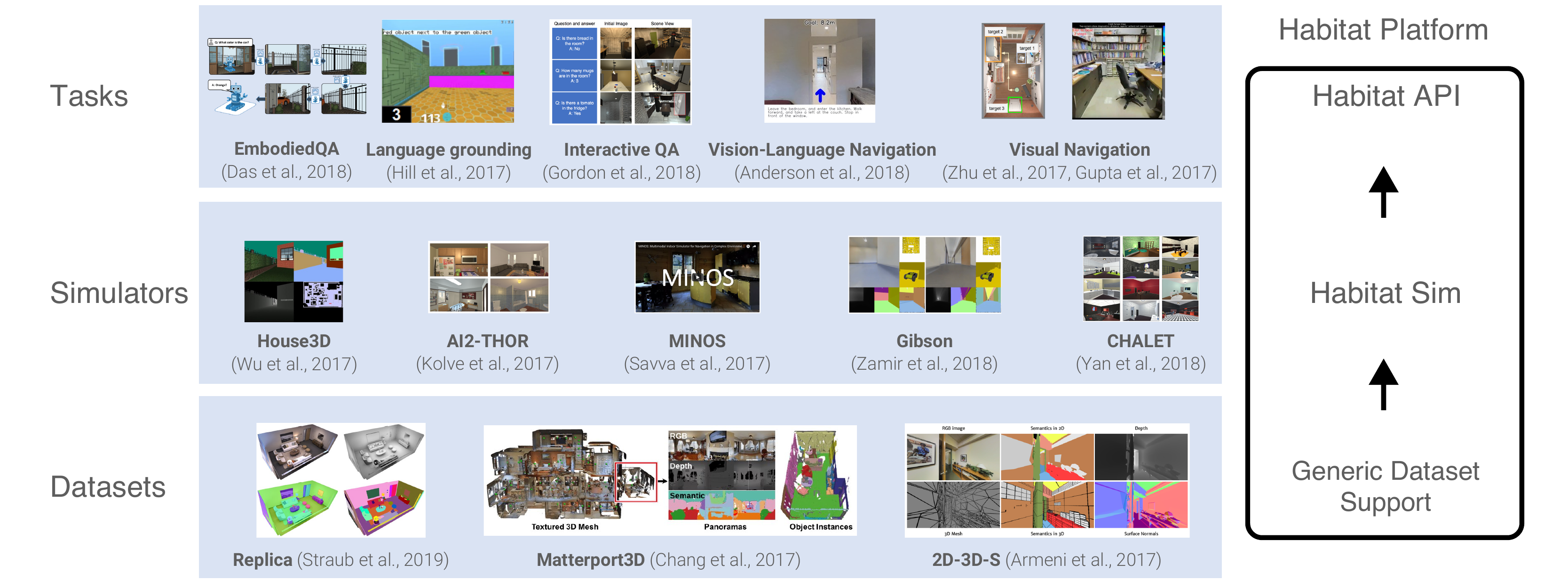}
    \caption{The `software stack' for training embodied agents involves 
    (1) \emph{datasets} providing 3D assets with semantic annotations, 
    (2) \emph{simulators} that render these assets and 
    within which an embodied agent may be simulated, and 
    (3) \emph{tasks} that define evaluatable problems that enable us to benchmark scientific progress.
    Prior work (highlighted in blue boxes) has contributed a variety of datasets, simulation software, and task definitions.
    We propose a unified embodied agent stack with the Habitat platform, including generic dataset support, a highly performant simulator (\habitatsim), and a flexible API (\habitatapi) allowing the definition and evaluation of a broad set of tasks.}
    \label{fig:habitat_stack}
\end{figure*}

We aim to support a complementary research program: training embodied agents (\eg virtual robots) in rich realistic simulators and then transferring the learned skills to reality.
Simulations have a long and rich history in science and engineering (from aerospace to zoology).
In the context of embodied AI, simulators help overcome the aforementioned challenges -- they can run orders of magnitude faster than real-time and can be parallelized over a cluster; training in simulation is safe, cheap, and enables fair comparison and benchmarking of progress in a concerted community-wide effort.
Once a promising approach has been developed and tested in simulation, it can be transferred to physical platforms that operate in the real world~\cite{sim2real-driving,sim2real-legged}.

Datasets have been a key driver of progress in computer vision, NLP, and other areas of AI~\cite{Deng09imagenet,mscoco,antol_iccv15,ammirato2017dataset}.
As the community transitions to embodied AI, we believe that simulators will 
assume the role played previously by datasets. 
To support this transition, 
we aim to standardize the entire `software stack' for training embodied agents 
(\Cref{fig:habitat_stack}): 
scanning the world and creating photorealistic 3D assets, 
developing the next generation of highly efficient and parallelizable simulators, 
specifying embodied AI tasks that enable us to benchmark scientific progress, 
and releasing modular high-level libraries for training and deploying embodied agents. 
%
Specifically, Habitat consists of the following: 
\begin{asparaenum}

\item \habitatsim: 
a flexible, high-performance 3D simulator with configurable agents, 
multiple sensors, and generic 3D dataset handling
(with built-in support for Matterport3D, Gibson, and Replica datasets). 

\item \habitatapi: a modular high-level library for 
end-to-end development of embodied AI algorithms -- 
defining embodied AI tasks (\eg navigation,
instruction following, question answering), 
configuring 
and training embodied agents (via imitation or reinforcement learning, 
or via classic SLAM), and 
benchmarking 
using standard metrics~\cite{Anderson2018-Evaluation}.

\end{asparaenum}

The Habitat architecture and implementation combine modularity and high performance.
When rendering a scene from the Matterport3D dataset, \habitatsim achieves several thousand frames per second (fps) running single-threaded, and can reach over $10@000$ fps multi-process on a single GPU, which is orders of magnitude faster than the closest simulator. 
\habitatapi allows us to train and benchmark embodied agents with different classes of methods and in different 3D scene datasets. 

These large-scale engineering contributions enable us to answer scientific questions requiring experiments that were till now impracticable or `merely' impractical.
Specifically, in the context of point-goal navigation~\cite{Anderson2018-Evaluation}, we 
make two scientific contributions: 
\begin{asparaenum}
\item We revisit the comparison between learning and SLAM approaches from two recent works~\cite{mishkin2019benchmarking, kojima2019learn}
and find evidence for the \textbf{opposite conclusion} -- that learning outperforms SLAM if scaled to an order of magnitude more experience than previous investigations.

\item We conduct the first cross-dataset generalization experiments $\{$train, test$\} \times \{$Matterport3D, Gibson$\}$ for multiple sensors $\{$Blind\footnote{$Blind$ refers to agents with no visual sensory inputs.}, RGB, RGBD, D$\} \times \{$GPS+Compass$\}$ and find that only agents with depth ($D$) sensors generalize well across datasets.
\end{asparaenum}

We hope that our open-source platform and these findings will advance and guide 
future research in embodied AI. 

\section{Related Work}

\setlength{\epigraphwidth}{0.55\columnwidth}
\epigraph{Reality is something you rise above.}{\textit{Liza Minnelli}}

The availability of large-scale 3D scene datasets~\cite{armeni_cvpr16,Song2017,Chang2017} and community interest in active vision tasks led to a recent surge of work that resulted in the development of a variety of simulation platforms for indoor environments~\cite{Kolve2017,Brodeur2017,gupta_cvpr17,Savva2017,Wu2018,Anderson2018-Language,Xia2018,Yan2018,Puig2018}.
These platforms vary with respect to the 3D scene data they use, the embodied agent tasks they address, and the evaluation protocols they implement.

This surge of activity is both thrilling and alarming. On the one hand, it is clearly a sign of the interest in embodied AI across diverse research communities (computer vision, natural language processing, robotics, machine learning).
On the other hand, the existence of multiple differing simulation environments can cause fragmentation, replication of effort, and difficulty in reproduction and community-wide progress.
Moreover, existing simulators exhibit several shortcomings:
\begin{compactitem}[\hspace{1pt}--]
    \item Tight coupling of task (\eg navigation), simulation platform (\eg GibsonEnv), and 3D dataset (\eg Gibson). 
    Experiments with multiple tasks or datasets are impractical. 
    \item Hard-coded agent configuration (\eg size, action-space). Ablations of agent parameters and sensor types are not supported, making results hard to compare.
    \item Suboptimal rendering and simulation performance. Most existing indoor simulators operate at relatively low frame rates (10-100 fps), becoming a bottleneck in training agents and 
    making large-scale learning infeasible. Takeaway messages from such experiments 
    become unreliable -- has the learning converged to trust the comparisons? 
    \item Limited control of environment state. The structure of the 3D scene in terms of present objects cannot be programmatically modified (\eg to test the robustness of agents). 
\end{compactitem}

Most critically, work built on top of any of the existing platforms is hard to reproduce independently from the platform, and thus hard to evaluate against work based on a different platform, even in cases where the target tasks and datasets are the same.
This status quo is undesirable and motivates the Habitat effort.
We aim to learn from the successes of previous frameworks and develop a unifying platform that combines their desirable characteristics while addressing their limitations.
A common, unifying platform can significantly accelerate research by enabling code re-use and consistent experimental methodology.
Moreover, a common platform enables us to easily carry out experiments testing agents based on different paradigms (learned vs. classical) and generalization of agents between datasets.

The experiments we carry out contrasting learned and classical approaches to navigation are similar to the recent work of Mishkin et al.~\cite{mishkin2019benchmarking}.
However, the performance of the Habitat stack relative to MINOS~\cite{Savva2017} used in 
\cite{mishkin2019benchmarking} -- thousands vs.~one hundred frames per second -- 
allows us to evaluate agents that have been trained with significantly larger amounts of 
experience (75 million steps vs.~five million steps). 
The trends we observe demonstrate that learned agents can begin to match and outperform classical approaches when provided with large amounts of training experience.
Other recent work by Koijima and Deng~\cite{kojima2019learn} has also compared hand-engineered navigation agents against learned agents but their focus is on defining additional metrics to characterize the performance of agents and to establish measures of hardness for navigation episodes.
To our knowledge, our experiments are the first to train navigation agents provided with multi-month experience in realistic indoor environments and contrast them against classical methods.

\section{Habitat Platform}

\setlength{\epigraphwidth}{0.6\columnwidth}
\epigraph{What I cannot create I do not understand.}{\textit{Richard Feynman}}

The development of Habitat is a long-term effort to 
enable the formation of a common task framework~\cite{Donoho2015} for research into embodied agents, thereby supporting systematic research progress in this area.

\begin{figure}[t]
  \centering
  \includegraphics[width=0.32\columnwidth]{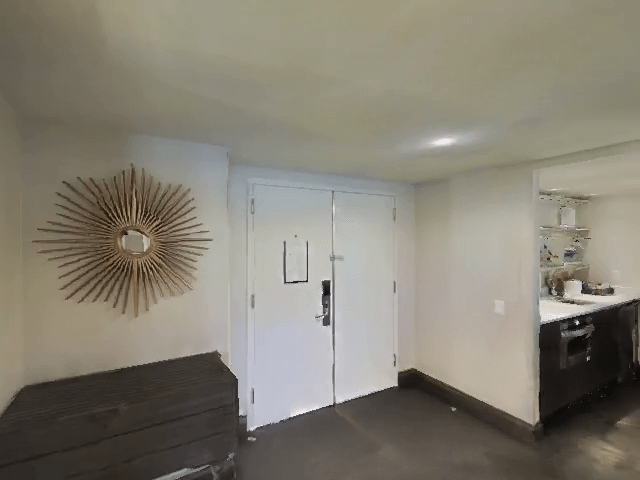}
  \includegraphics[width=0.32\columnwidth]{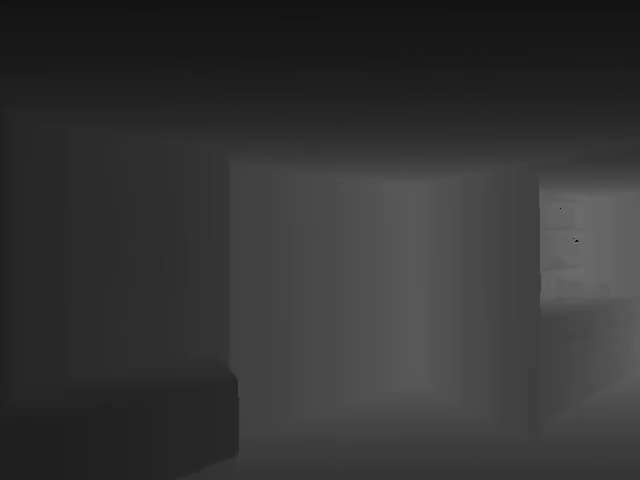}
  \includegraphics[width=0.32\columnwidth]{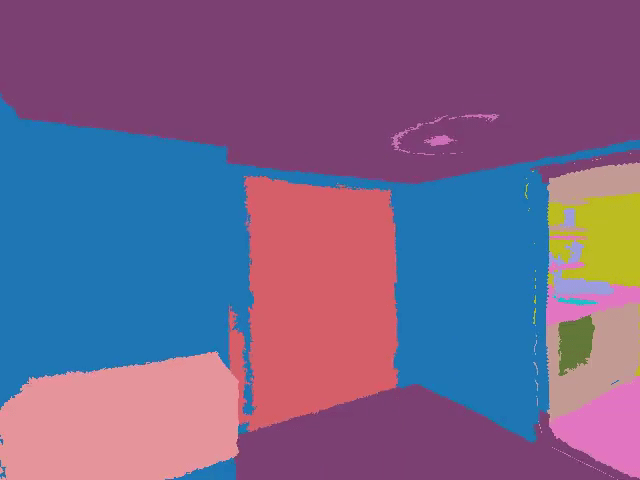}\\
  \includegraphics[width=0.32\columnwidth]{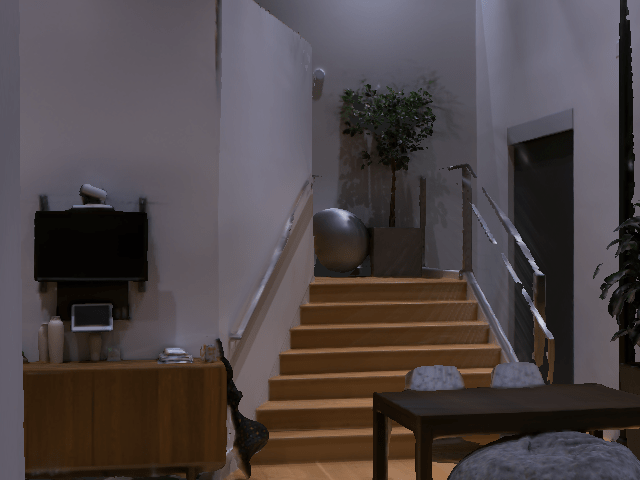}
  \includegraphics[width=0.32\columnwidth]{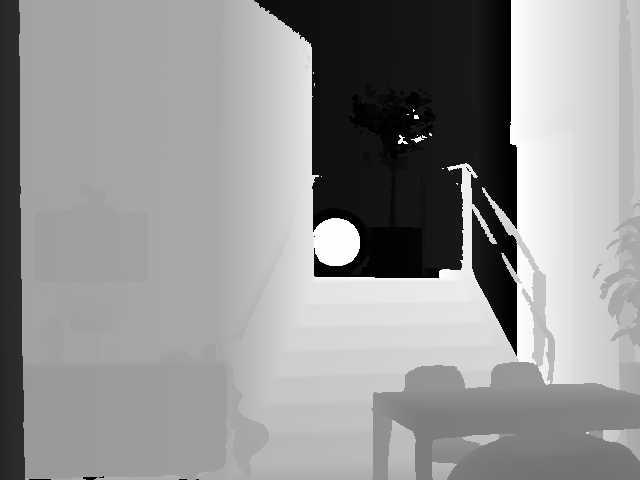}
  \includegraphics[width=0.32\columnwidth]{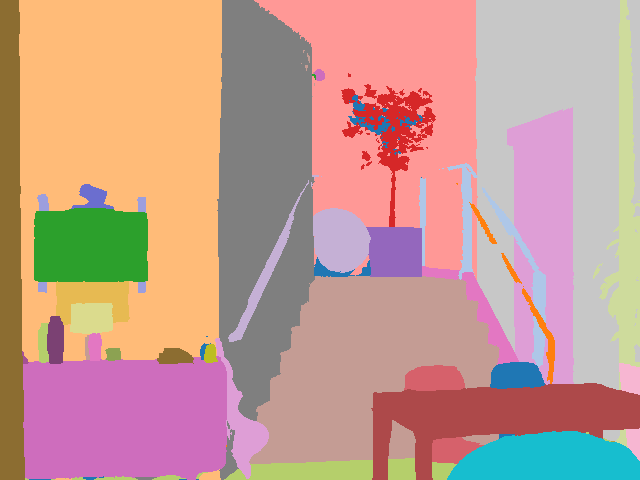}\\
  \caption{Example rendered sensor observations for three sensors (color camera, depth sensor, semantic instance mask) in two different environment datasets. A Matterport3D~\cite{Chang2017} environment is in the top row, and a Replica~\cite{replica19arxiv} environment in the bottom row.}
\label{fig:frames}
\end{figure}

\xhdr{Design requirements.}
The issues discussed in the previous section lead us to a set of requirements that we seek to fulfill. 

\begin{compactitem}[\hspace{1pt}--]
    \item \textbf{Highly performant rendering engine}: resource-efficient rendering engine that can produce multiple channels of visual information (\eg RGB, depth, semantic instance segmentation, 
    surface normals, optical flow) for multiple concurrently operating agents.
    \item \textbf{Scene dataset ingestion API}: makes the platform agnostic to 3D scene datasets and allows users to use their own datasets. 
    \item \textbf{Agent API}: allows users to specify parameterized embodied agents with well-defined geometry, physics, and actuation characteristics. 
    \item \textbf{Sensor suite API}: allows specification of arbitrary numbers of 
    parameterized sensors (\eg RGB, depth, contact, GPS, compass sensors) attached 
    to each agent. 
    \item \textbf{Scenario and task API}: allows portable definition of tasks and their evaluation protocols.
    \item \textbf{Implementation}: C++ backend with Python API and interoperation with common learning frameworks, minimizes entry threshold.
    \item \textbf{Containerization}: enables distributed training in clusters and remote-server evaluation of user-provided code.
    \item \textbf{Humans-as-agents}: allows humans to function as agents in simulation in order to collect human behavior and investigate human-agent or human-human interactions. 
    \item \textbf{Environment state manipulation}: programmatic control of the environment configuration in terms of the objects that are present and their relative layout. 
\end{compactitem}

\xhdr{Design overview.}
The above design requirements cut across several layers in the `software stack' in \Cref{fig:habitat_stack}.
A monolithic design is not suitable for addressing requirements at all levels. We, therefore, structure the Habitat platform to mirror this multi-layer abstraction. 

At the lowest level is
\habitatsim, a flexible, high-performance 3D simulator, 
responsible for 
loading 3D scenes into a standardized scene-graph representation, 
configuring agents with multiple sensors, simulating agent motion, and returning sensory data from an agent's sensor suite.
The sensor abstraction in Habitat allows additional sensors such as LIDAR and IMU to be easily implemented as plugins.

\xhdr{Generic 3D dataset API using scene graphs.}
\habitatsim employs a hierarchical scene graph to 
represent all supported 3D environment datasets, 
whether synthetic or based on real-world reconstructions.
The use of a uniform scene graph representation allows us to abstract the details of specific datasets, and to treat them in a consistent fashion.
Scene graphs allow us to compose 3D environments through procedural scene generation, editing, or programmatic manipulation.

\xhdr{Rendering engine.}
The \habitatsim backend module is implemented in C++ and leverages the Magnum graphics middleware library\footnote{\url{https://magnum.graphics/}} to support cross-platform deployment on a broad variety of hardware configurations.
The simulator backend employs an efficient rendering pipeline that implements visual sensor frame rendering using a multi-attachment `uber-shader' combining outputs for color camera sensors, depth sensors, and semantic mask sensors.
By allowing all outputs to be produced in a single render pass, we avoid additional overhead when sensor parameters are shared and the same render pass can be used for all outputs.
\Cref{fig:frames} shows examples of visual sensors rendered in three different supported datasets.
The same agent and sensor configuration was instantiated in a scene from each of the three datasets by simply specifying a different input scene.


\xhdr{Performance.}
\habitatsim achieves thousands of frames per second per simulator thread and is orders of magnitude faster than previous simulators for realistic indoor environments (which typically operate at tens or hundreds of frames per second) -- see \Cref{tab:performance_summary} for a summary and the supplement for more details.
By comparison, AI2-THOR~\cite{Kolve2017} and CHALET~\cite{Yan2018} run at tens of fps, MINOS~\cite{Savva2017} and Gibson~\cite{Xia2018} run at about a hundred, and House3D~\cite{Wu2018} runs at about $300$ fps. \habitatsim is $2$-$3$ orders of magnitude faster.
%
By operating at $10@000$ frames per second we shift the bottleneck from simulation to optimization for network training.
Based on TensorFlow benchmarks, many popular network architectures run at frame rates that are 10-100x lower on a single GPU\footnote{\url{https://www.tensorflow.org/guide/performance/benchmarks}}.
In practice, we have observed that it is often \emph{faster to generate images using \habitatsim than to load images from disk}.

\begin{table}[t!]
  \centering
\resizebox{\columnwidth}{!}{
  \begin{tabular}{@{}lrrrrrr@{}}

      \toprule
      & \multicolumn{3}{c}{$1$ process} 
      & \multicolumn{3}{c}{$5$ processes}\\
      \cmidrule(l){2-4} 
      \cmidrule(l){5-7} 
      Sensors / Resolution 
      & $128$ & $256$ & $512$ 
      & $128$ & $256$ & $512$\\
      \midrule
      RGB 
      & $4@093$ & $1@987$ & $848$ 
      & $10@592$ & $3@574$ & $2@629$\\
      RGB + depth 
      & $2@050$ & $1@042$ & $423$ 
      & $5@223$ & $1@774$ & $1@348$\\
      \bottomrule
  \end{tabular}
}
  \caption{Performance of \habitatsim in frames per second for an example Matterport3D scene (id 17DRP5sb8fy) on an Intel Xeon E5-2690 v4 CPU and Nvidia Titan Xp GPU, measured at different frame resolutions and with a varying number of concurrent simulator processes sharing the GPU. See the supplement for additional benchmarking results.}
  \label{tab:performance_summary}
\end{table}

\xhdr{Efficient GPU throughput.}
Currently, frames rendered by \habitatsim are exposed as Python tensors through shared memory. Future development will focus on even higher rendering efficiency by entirely avoiding GPU-to-CPU memory copy overhead through the use of CUDA-GL interoperation and direct sharing of render buffers and textures as tensors. 
Our preliminary internal testing suggests that this can lead to 
a speedup by a factor of 2. 

Above the simulation backend, the \habitatapi layer is  a modular high-level library for end-to-end development in embodied AI.
%
%
%
Setting up an embodied task involves 
specifying observations that may be used by the agent(s), 
using environment information provided by the simulator, and 
connecting the information with a task-specific episode dataset.  

\begin{compactitem}[\hspace{1pt}--]
    

    \item \texttt{Task}: this class extends the simulator's \texttt{Observations} class and action space with task-specific ones. The criteria of episode termination and measures of success are provided by the \texttt{Task}. For example, 
    in goal-driven navigation, \texttt{Task} provides 
    the goal and evaluation metric~\cite{Anderson2018-Evaluation}. 
    To support this kind of functionality the \texttt{Task} has read-only access to \texttt{Simulator} and \texttt{Episode-Dataset}.

    \item \texttt{Episode}: a class for episode specification that includes the initial position and orientation of an \texttt{Agent}, scene id,  goal position, and optionally the shortest path to the goal. An episode is a description of an instance of the task. 
    
    
    \item \texttt{Environment}: the fundamental environment concept for Habitat, abstracting all the information needed for working on embodied tasks with a simulator.  

\end{compactitem}



More details about the architecture of the Habitat platform, performance measurements, and examples of API use are provided in the supplement.

\section{PointGoal Navigation at Scale}

To demonstrate the utility of the Habitat platform design, we carry out experiments to test for generalization of goal-directed visual navigation agents between datasets of different environments and to compare the performance of learning-based agents against classic agents as the amount of available training experience is increased.

\xhdr{Task definition.}
We use the PointGoal task (as defined by Anderson~\etal~\cite{Anderson2018-Evaluation}) as our 
experimental testbed. This task is ostensibly simple to define -- 
an agent is initialized at a random starting position and 
orientation in an environment and asked to navigate to target coordinates that are provided relative to the agent's position; 
no ground-truth map is available and the agent must only use
its sensory input to navigate. 
However, in the course of experiments, we realized that 
this 
task leaves space for
subtle choices that (a) can make a significant difference in 
experimental outcomes and (b) are either not specified 
or inconsistent across papers, making comparison difficult. 
We attempt to be as descriptive as possible about these 
seemingly low-level choices; we hope the Habitat platform will help iron out these inconsistencies. 

\xhdr{Agent embodiment and action space.}
The agent is physically embodied as a cylindrical primitive shape with diameter $0.2\text{m}$ and height $1.5\text{m}$.
The action space consists of four actions: \turnleft, \turnright, \forward, and \callstop.
These actions are mapped to idealized actuations that result in $10$ degree turns for the turning actions and linear displacement of $0.25\text{m}$ for the \forward action. 
The \callstop action allows the agent to signal that it has reached the goal.
Habitat supports noisy actuations but experiments in this paper are conducted in the noise-free setting as our analysis focuses on other factors.

\xhdr{Collision dynamics.} 
Some previous works~\cite{Anderson2018-Language} use a coarse 
irregular navigation graph where an agent effectively 
`teleports' from one location to another ($1$-$2\text{m}$ apart). Others~\cite{embodiedqa} use a fine-grained regular grid 
($0.01\text{m}$ resolution) where the agent moves on unoccupied cells 
and there are no collisions or partial steps. 
In Habitat and our experiments, we use a more realistic 
collision model -- 
the agent navigates in 
a continuous state space\footnote{Up to machine precision.}
and motion can produce collisions resulting in partial 
(or no) progress along the direction intended -- simply put, 
it is possible for the agent to `slide' along a wall or obstacle. 
Crucially, the agent may choose \forward (0.25m) and 
end up in a location that is \emph{not} 0.25m 
forward of where it started; thus, 
odometry is not trivial even in the absence of actuation noise. 

\xhdr{Goal specification: static or dynamic?}
One conspicuous underspecification in the PointGoal task~\cite{Anderson2018-Evaluation} is whether the goal coordinates 
are \emph{static} (\ie provided once at the start of the episode) 
or \emph{dynamic} (\ie provided at \emph{every} time step). 
The former is more realistic -- it is difficult to imagine a real 
task where an oracle would provide precise dynamic goal coordinates. 
However, in the absence of actuation noise and collisions, 
every step taken by the agent results in a known turn or 
translation, and this combined with the initial goal location is 
functionally equivalent to dynamic goal specification. 
We hypothesize that this is why recent works~\cite{kojima2019learn, mishkin2019benchmarking, gupta_cvpr17} used dynamic 
goal specification. 
We follow and prescribe the following conceptual delineation 
-- as a \emph{task}, we adopt static PointGoal navigation; 
as for the \emph{sensor suite}, we equip our 
agents with an idealized GPS+Compass sensor.  
This orients us towards a realistic task (static PointGoal navigation), disentangles simulator design 
(actuation noise, collision dynamics) from the task definition, 
and allows us to compare techniques by sensors used 
(RGB, depth, GPS, compass, contact sensors).

\xhdr{Sensory input.}
The agents are endowed with a single color vision sensor placed at a height of $1.5\text{m}$ from the center of the agent's base and oriented to face `forward'.
This sensor provides RGB frames at a resolution of $256^2$ pixels 
and with a field of view of $90$ degrees.
In addition, an idealized depth sensor is available, in the same position and orientation as the color vision sensor.
The field of view and resolution of the depth sensor 
match those of the color vision sensor.
We designate agents that make use of the color sensor by \rgb, 
agents that make use of the depth sensor by \depth, 
and agents that make use of both by \rgbd.
Agents that use neither sensor are denoted as \blind.
All agents are equipped with an idealized GPS and compass 
-- \ie, they have access to their location coordinates, and 
implicitly their orientation relative to the goal position.

\xhdr{Episode specification.}
We initialize the agent at a starting position and 
orientation that are sampled uniformly at random from all 
navigable positions on the floor of the environment. 
The goal position is chosen such that it lies on the 
same floor and there exists a navigable path from the 
agent's starting position. 
During the episode, the agent is allowed to take up to 
\maxsteps actions.
This threshold significantly exceeds the number of steps an optimal agent requires to reach all goals (see the supplement).
After each action, the agent receives a set of 
observations from the active sensors. 

\xhdr{Evaluation.}
A navigation episode is considered successful if and only if the agent issues a \callstop action within $0.2\text{m}$ of the target coordinates, as measured by a geodesic distance along the shortest path from the agent's position to the goal position.
If the agent takes \maxsteps actions without the above condition being met the episode ends and is considered unsuccessful.
Performance is measured using the `Success weighted by Path Length' (\spl) metric~\cite{Anderson2018-Evaluation}.
For an episode where the geodesic distance of the 
shortest path is $l$ and the agent traverses a distance $p$, 
\spl is defined as $S \cdot \nicefrac{l}{\max(p,l)}$, where $S$ is a binary indicator of success.

\xhdr{Episode dataset preparation.}
We create PointGoal navigation episode-datasets for Matterport3D~\cite{Chang2017} and Gibson~\cite{Xia2018} scenes. 
For Matterport3D we followed the publicly available train/val/test splits.
Note that as in recent works~\cite{embodiedqa, mishkin2019benchmarking, kojima2019learn}, 
there is no overlap between train, val, and test scenes. 
For Gibson scenes, we obtained textured 3D surface meshes from the Gibson authors~\cite{Xia2018}, manually annotated each scene 
on its reconstruction quality 
(small/big holes, floating/irregular surfaces, poor textures), 
and curated a subset of 106 scenes (out of 572); 
see the supplement for details.
An episode is defined by the unique id of the scene, the starting position and orientation of the agent, and the goal position.
Additional metadata such as the geodesic distance along the shortest path (\gdsp) from start position to goal position is also included.
While generating episodes, we restrict the \gdsp 
to be between $1\text{m}$ and $30\text{m}$.
An episode is trivial if there is an obstacle-free straight line between the start and goal positions.
A good measure of the navigation complexity of an episode is the ratio of \gdsp to Euclidean distance between start and goal positions (notice that \gdsp can only be larger than or equal to the Euclidean distance).
If the ratio is nearly $1$, there are few obstacles and the episode is easy; if the ratio is much larger than $1$, the 
episode is difficult because strategic navigation is required. 
To keep the navigation complexity of the precomputed episodes reasonably high, we perform rejection sampling for episodes with the above ratio falling in the range $[1, 1.1]$. 
Following this, there is a significant decrease in the number of  near-straight-line episodes (episodes with a ratio in $[1, 1.1]$)
 -- from $37\%$ to $10\%$ for the Gibson dataset generation. 
This step was not performed in any previous studies.
We find that without this filtering, all metrics appear inflated.
Gibson scenes have smaller physical dimensions compared to 
the Matterport3D scenes.
This is reflected in the resulting PointGoal dataset -- 
average \gdsp of episodes in Gibson scenes is smaller than that of Matterport3D scenes.

\begin{figure*}
    \centering
    \includegraphics[width=\columnwidth]{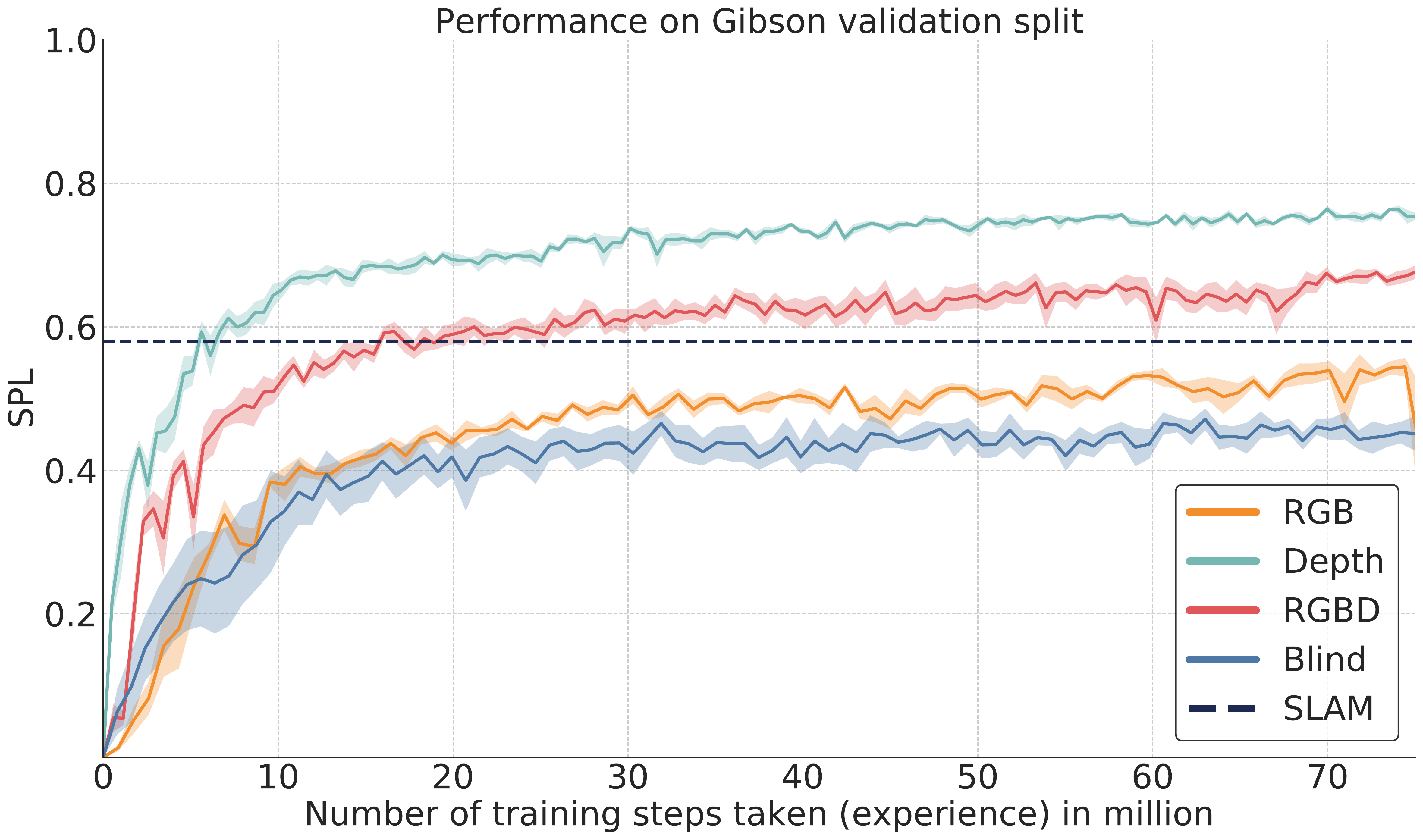}\quad\quad
    \includegraphics[width=\columnwidth]{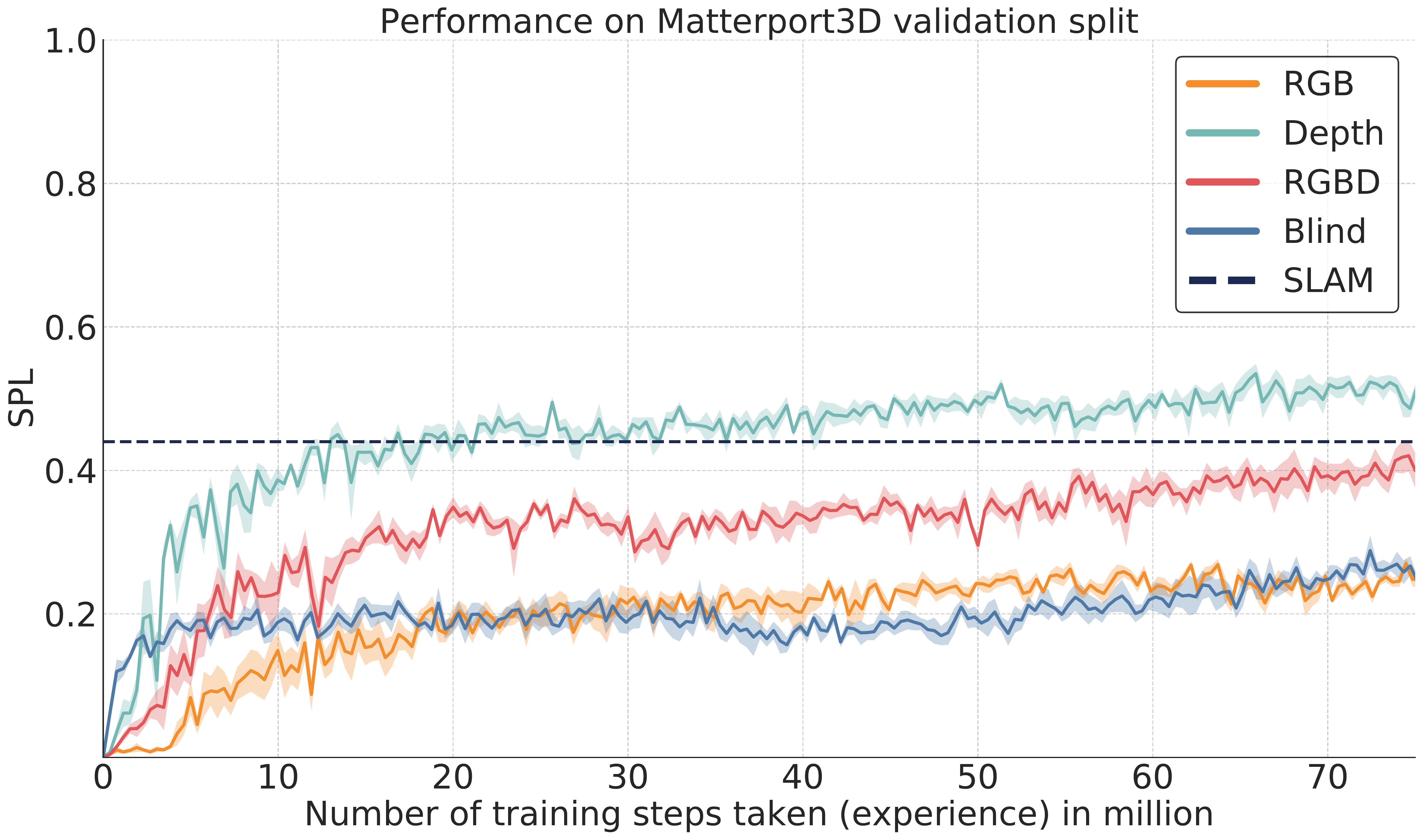}
    \caption{Average SPL of agents on the val set over the course of training. Previous work~\cite{mishkin2019benchmarking, kojima2019learn} has analyzed performance at 5-10 million steps. Interesting trends emerge with more experience: i)~\blind agents initially outperform \rgb and \rgbd but saturate quickly; ii)~Learning-based \depth agents outperform classic SLAM. The shaded areas around curves show the standard error of SPL over five seeds.
    }
    \label{fig:classical_learning}
\end{figure*}

\xhdr{Baselines.}
We compare the following baselines:
\begin{compactitem}[\hspace{1pt}--]
\item \textbf{Random} chooses an action randomly among \turnleft, \turnright, and \forward with uniform distribution.
The agent calls the \callstop action when within $0.2\text{m}$ of the goal 
(computed using the difference of static goal and dynamic GPS
coordinates).

\item \textbf{Forward only} always calls the \forward action, and calls the \callstop action when within $0.2\text{m}$ of the goal.

\item \textbf{Goal follower} moves towards the goal direction. If it is not facing the goal (more than $15$ degrees off-axis), it performs \turnleft or \turnright to align itself; otherwise, it calls \forward.
The agent calls the \callstop action when within $0.2\text{m}$ of the goal.

\item \textbf{RL (PPO)} is an agent trained with reinforcement 
learning, 
specifically proximal policy optimization \cite{Schulman2017ProximalPO}. 
We experiment with RL agents equipped with different 
visual sensors:  
no visual input (\blind), \rgb input, \depth input, and  RGB with depth (\rgbd). 
The model consists of a CNN that produces an embedding for visual input, which together with the relative goal vector 
is used by an actor (GRU) and a critic (linear layer).
The CNN has the following architecture: $\{$Conv 8$\times$8, ReLU, Conv 4$\times$4, ReLU, Conv  3$\times$3, ReLU, Linear, ReLU$\}$ (see supplement for details). 
Let $r_t$ denote the reward at timestep $t$, 
$d_{t}$ be the geodesic distance to goal at timestep $t$, 
$s$ a success reward and $\lambda$ a time penalty (to encourage 
efficiency). 
All models were trained with the following reward function:
$$
    r_t=\begin{cases}
s + d_{t - 1} - d_{t} + \lambda & \text{if goal is reached}
\\ d_{t - 1} - d_{t} + \lambda & \text{otherwise}
\end{cases}
$$
In our experiments $s$ is set to $10$ and $\lambda$ is set to $-0.01$.
Note that rewards are only provided in training environments; the task is challenging as the agent must generalize to unseen test environments.

\item \textbf{SLAM \cite{mishkin2019benchmarking}} 
is an agent implementing a classic robotics navigation pipeline 
(including components for localization, mapping, and planning), using RGB and depth sensors.
We use the classic agent by  Mishkin~\etal~\cite{mishkin2019benchmarking} which leverages the ORB-SLAM2~\cite{murORB2} localization pipeline, with the same parameters as reported in the original work.
\end{compactitem}

\vspace{1em}
\xhdr{Training procedure.}
When training learning-based agents, we first divide 
the scenes in the training set equally among 
$8$ (Gibson), $6$ (Matterport3D) concurrently running simulator worker threads. 
Each thread establishes blocks of $500$ training 
episodes for each scene in its training set partition 
and shuffles the ordering of these blocks. 
Training continues through shuffled copies of this array. 
We do not hardcode the \callstop action to retain generality and allow for comparison with future work that does not assume GPS inputs.
For the experiments reported here, we train until 
$75$ million agent steps are accumulated across 
all worker threads. This is 15x larger than the 
experience used in previous investigations~\cite{mishkin2019benchmarking, kojima2019learn}. 
Training agents to $75$ million steps took (in sum over all three datasets): $320$ GPU-hours for \blind, $566$ GPU-hours for \rgb, $475$ GPU-hours for \depth, and $906$ GPU-hours for \rgbd (overall $2267$ GPU-hours).

\section{Results and Findings}

\begin{table}
  \centering
  \begin{tabular}{@{} p{0.5in} p{0.85in} cccccc@{}}
      \toprule
      & & \multicolumn{2}{c}{Gibson} & \multicolumn{2}{c}{MP3D}\\
      \cmidrule(l){3-4} \cmidrule(l){5-6}
      Sensors & Baseline & SPL & Succ & SPL & Succ \\
      \midrule
      \multirow{4}{*}{\blind}
      & Random & $0.02$ & $0.03$ & $0.01$ & $0.01$ \\
      & Forward only & $0.00$ & $0.00$ & $0.00$ & $0.00$ \\
      & Goal follower & $0.23$ & $0.23$ & $0.12$ & $0.12$ \\
      & RL (PPO) & $0.42$ & $0.62$ & $0.25$ & $0.35$ \\
      \midrule
      \multirow{1}{*}{\rgb}
      & RL (PPO) & $0.46$ & $0.64$ & $0.30$ & $0.42$ \\
      \midrule
      \multirow{1}{*}{\depth}
      & RL (PPO) & $\textbf{0.79}$ & $\textbf{0.89}$ & $\textbf{0.54}$ & $\textbf{0.69}$ \\
      \midrule
      \multirow{2}{*}{\rgbd}
      & RL (PPO) & $0.70$ & $0.80$ & $0.42$ & $0.53$ \\
      & SLAM~\cite{mishkin2019benchmarking} & $0.51$ & $0.62$ & $0.39$ & $0.47$ \\
      \bottomrule
  \end{tabular}
  \caption{Performance of baseline methods on the PointGoal task~\cite{Anderson2018-Evaluation} tested on the Gibson~\cite{Xia2018} and MP3D~\cite{Chang2017} test sets under multiple sensor configurations. RL models have been trained for $75$ million steps. We report average rate of episode success and SPL~\cite{Anderson2018-Evaluation}.
  }
  \label{tab:baselines}
\end{table}


We seek to answer two questions: 
i) how do learning-based agents compare to classic SLAM 
and hand-coded baselines as the amount of training experience increases and 
ii) how well do learned agents generalize 
across 3D datasets. 

It should be tacitly understood, but to be explicit -- 
`learning' and `SLAM' are broad families of techniques 
(and not a single method), 
are not necessarily mutually exclusive, and are 
not `settled' in their development. We compare representative instances of these families to gain insight into questions of 
scaling and generalization, and do not make any claims 
about intrinsic superiority of one or the other.

\xhdr{Learning vs SLAM.}
To answer the first question we plot agent performance 
(\spl) on validation (\ie unseen) episodes over the 
course of training in \Cref{fig:classical_learning}
(top: Gibson, bottom: Matterport3D). 
SLAM~\cite{mishkin2019benchmarking} 
does not require training 
and thus has a constant performance (0.59 on Gibson, 0.42 on 
Matterport3D). 
All RL (PPO) agents start out with far worse \spl, 
but RL (PPO) \depth, in particular, improves dramatically 
and matches the classic baseline 
at approximately $10$M frames 
(Gibson) or $30$M frames (Matterport3D)  
of experience, continuing to improve thereafter.  
Notice that if we terminated the experiment at $5$M frames 
as in \cite{mishkin2019benchmarking} we would 
also conclude that SLAM~\cite{mishkin2019benchmarking} dominates. 
Interestingly, 
\rgb agents do not
significantly outperform \blind agents; we hypothesize 
because both 
are equipped with GPS sensors. Indeed, qualitative 
results (\Cref{fig:top_down_nav} and video in supplement) 
suggest that \blind agents `hug' walls and 
implement `wall following' heuristics.
In contrast, \rgb sensors provide a high-dimensional complex 
signal that may be prone to overfitting to train environments 
due to the variety across scenes (even within the same dataset). 
We also notice in \Cref{fig:classical_learning} that \emph{all methods} perform better on Gibson than Matterport3D. 
This is consistent with our previous analysis that Gibson 
contains smaller scenes and shorter episodes. 

Next, for each agent and dataset, we select the best-performing checkpoint on validation and report results on test in \Cref{tab:baselines}. 
We observe that uniformly across the datasets, 
RL (PPO) \depth performs best, 
outperforming RL (PPO) \rgbd (by 0.09-0.16 \spl), 
SLAM (by 0.15-0.28 \spl), and \rgb (by 0.13-0.33 \spl) in that order (see the supplement for additional experiments involving noisy depth).
We believe \depth performs better than \rgbd because i) the PointGoal navigation task requires reasoning only about free space and depth provides relevant information directly, and ii) \rgb has significantly more entropy (different houses look very different), thus it is easier to overfit when using \rgb.
We ran our experiments with 5 random seeds per run, to confirm that these differences are statistically significant.
The differences are about an order of magnitude larger than the standard deviation of average SPL for all cases (\eg on the Gibson dataset errors are, \depth: $\pm 0.015$, \rgb: $\pm 0.055$, \rgbd: $\pm 0.028$, \blind: $\pm 0.005$).
Random and forward-only agents have very low performance, while the hand-coded goal follower and \blind baseline see modest performance.
See the supplement for additional analysis of trained agent behavior.

\begin{figure}
    \centering
    \includegraphics[width=\linewidth]{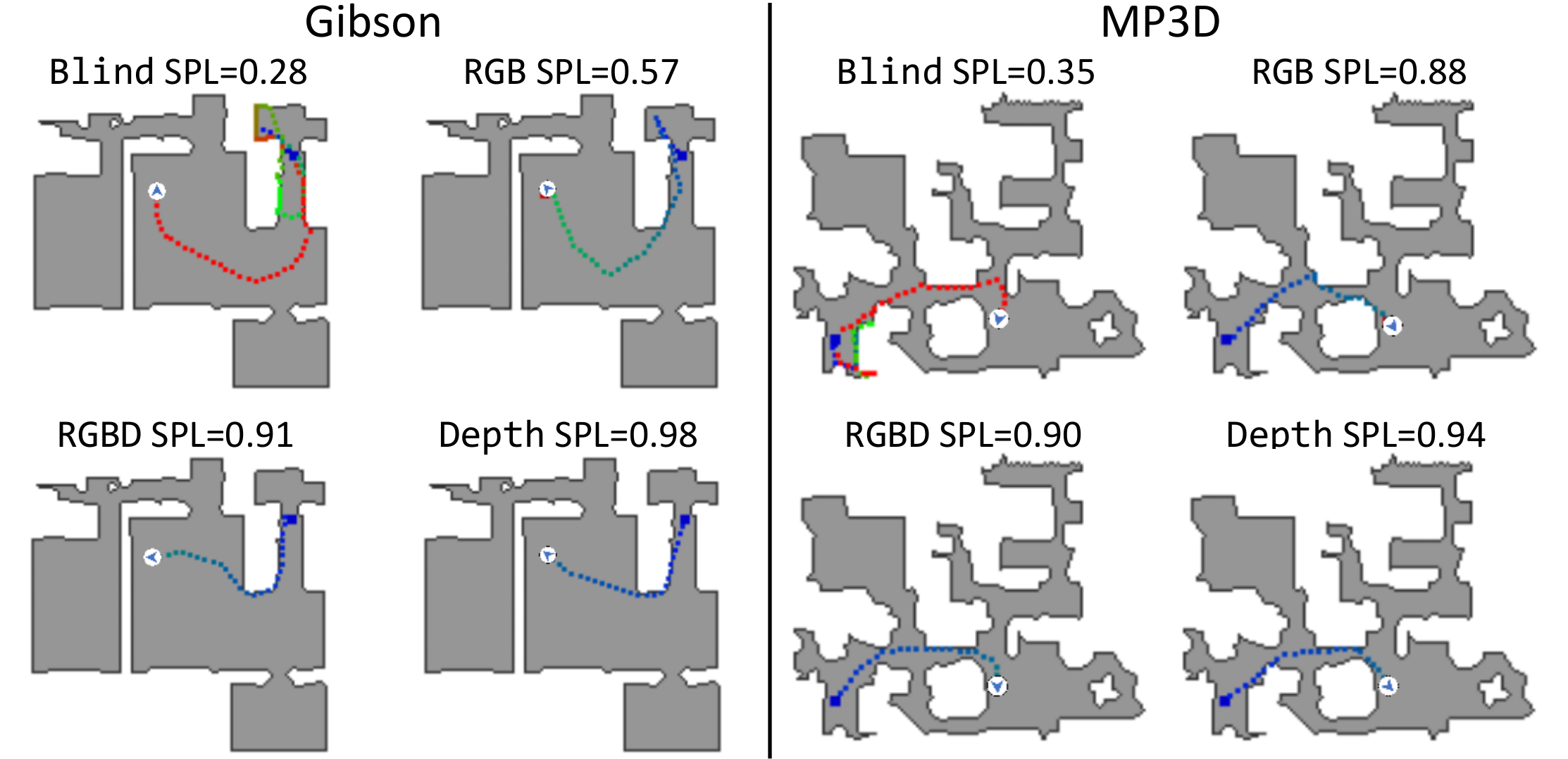}
    \caption{Navigation examples for different sensory configurations of the RL (PPO) agent, visualizing trials from the Gibson and MP3D val sets. A \textbf{blue dot} and \textbf{red dot} indicate the starting and goal positions, and the \textbf{blue arrow} indicates final agent position. The \textbf{blue-green-red line} is the agent's trajectory. Color shifts from blue to red as the maximum number of agent steps is approached.
    See the supplemental materials for more example trajectories.
    }
    \label{fig:top_down_nav}
\end{figure}

In \Cref{fig:top_down_nav} we plot example trajectories for 
the RL (PPO) agents, to qualitatively contrast their 
behavior in the same episode. 
Consistent with the aggregate statistics, we observe that \blind collides with obstacles and follows walls, while \depth is 
the most efficient.
See the supplement and the video for more example trajectories.

\begin{figure}
    \centering
    \includegraphics[width=\columnwidth]{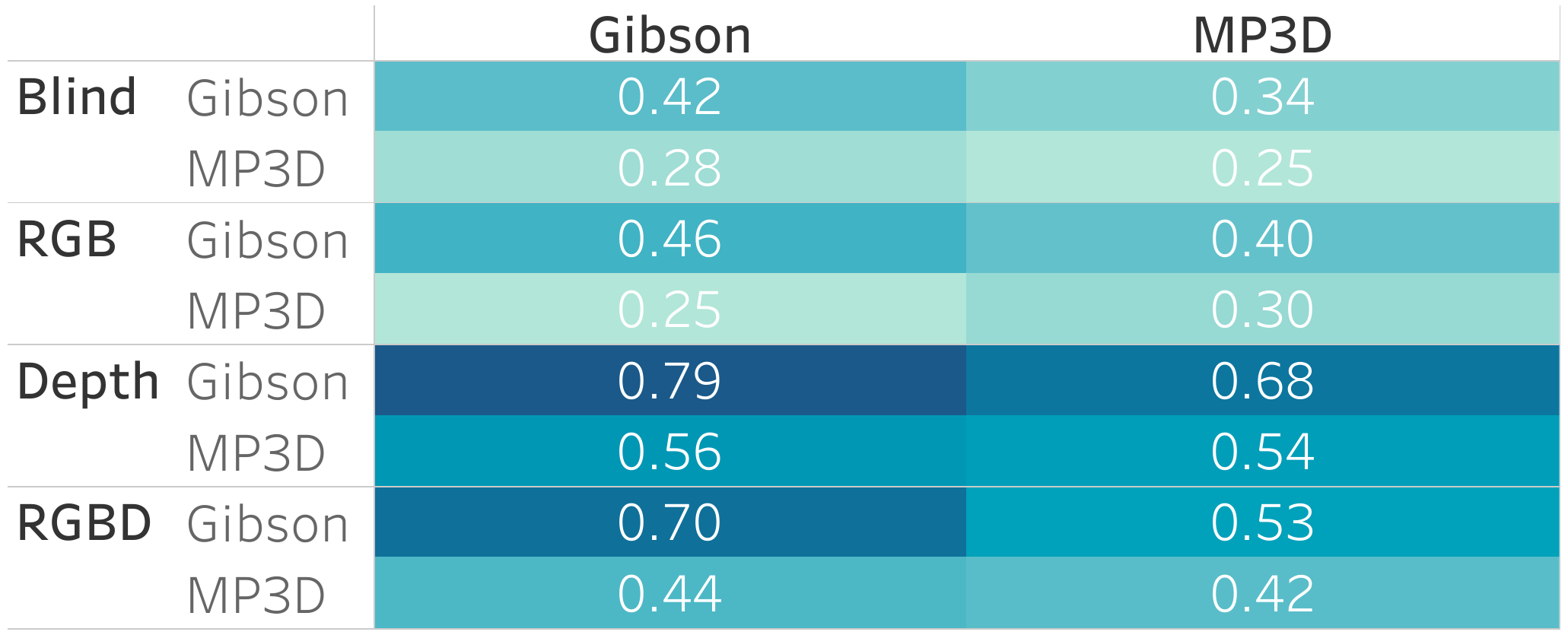}
    \caption{Generalization of agents between datasets. We report average SPL for a model trained on the source dataset in each row, as evaluated on test episodes for the target dataset in each column.}
    \label{fig:generalization}
\end{figure}

\xhdr{Generalization across datasets.}
Our findings so far are that RL (PPO) agents significantly outperform SLAM~\cite{mishkin2019benchmarking}. This prompts our second question
-- are these findings dataset specific or do learned agents generalize across datasets?
We report exhaustive comparisons in \Cref{fig:generalization} -- 
specifically, average \spl for all combinations of 
$\{$train, test$\} \times \{$Matterport3D, Gibson$\}$ for 
all agents $\{$\blind, \rgb, \rgbd, \depth$\}$. 
Rows indicate (agent, train set) pair, columns indicate test set. 
We find a number of interesting trends. 
First, nearly all agents suffer a drop in performance 
when trained on one dataset and tested on another, 
\eg \rgbd Gibson$\rightarrow$Gibson $0.70$ vs \rgbd 
Gibson$\rightarrow$Matterport3D $0.53$ (drop of 0.17). 
\rgb and \rgbd agents suffer a 
significant performance degradation, 
while the \blind agent is least affected (as we would expect).


Second, we find a potentially counter-intuitive trend -- 
agents trained on Gibson consistently outperform their 
counterparts trained on Matterport3D, \emph{even 
when evaluated on Matterport3D}.
We believe the reason is the previously noted observation that 
Gibson scenes are smaller and episodes are shorter 
(lower \gdsp) than Matterport3D. 
Gibson agents are trained on `easier' episodes and 
encounter positive reward more easily during random exploration, 
thus bootstrapping learning. Consequently, for a fixed 
computation budget Gibson agents are stronger universally (not 
just on Gibson). 
This finding suggests that visual navigation agents could 
benefit from curriculum learning. 

These insights are enabled by the engineering of Habitat, which made these experiments as simple as a change in the evaluation dataset name.
\section{Habitat Challenge}

\setlength{\epigraphwidth}{0.8\columnwidth}
\epigraph{No battle plan ever survives contact with the enemy.}{\textit{Helmuth Karl Bernhard von Moltke}}

Challenges drive progress. The history of AI sub-fields indicates that the formulation of the right questions, the creation of the right datasets, and the coalescence of communities around the right challenges drives scientific progress. 
Our goal is to support this process for embodied AI. 
\habitatchal is an autonomous navigation challenge  
that aims to benchmark and advance efforts in 
goal-directed visual navigation. 

One difficulty in creating a challenge 
around embodied AI tasks is the transition from 
static predictions (as in passive perception) to sequential 
decision making (as in sensorimotor control).
In traditional `internet AI' challenges 
(\eg ImageNet~\cite{Deng09imagenet}, COCO~\cite{mscoco}, VQA~\cite{antol_iccv15}), it is possible to release a static testing 
dataset and ask participants to simply upload their predictions on this 
set. In contrast, embodied AI tasks typically involve sequential decision making and agent-driven control, making it infeasible to pre-package a testing dataset. 
Essentially, embodied AI challenges require participants 
to \emph{upload code not predictions}. The uploaded agents 
can then be evaluated in novel (unseen) test environments. 

\xhdr{Challenge infrastructure.}
We leverage the frontend and challenge submission process 
of the EvalAI platform, and build backend infrastructure ourselves. 
Participants in \habitatchal are asked to 
upload Docker containers~\cite{merkel2014docker} with their 
agents via EvalAI. The submitted agents are then evaluated on a live AWS GPU-enabled instance. 
Specifically, contestants are free to train their agents however they wish 
(any language, any framework, any infrastructure). 
In order to evaluate these agents, participants are asked to 
derive from a base Habitat Docker container and 
implement a specific interface to their model -- 
agent's action taken given an observation 
from the environment at each step. This dockerized interface enables running the participant code on new environments. 

More details regarding the \habitatchal held at CVPR 2019 are available at the \url{https://aihabitat.org/challenge/} website.
In a future iteration of this challenge we will introduce three major differences designed to both reduce the gap between simulation and reality and to increase the difficulty of the task.
\begin{compactitem}[--]
  \item In the 2019 challenge, the relative coordinates specifying the goal were continuously updated during agent movement -- essentially simulating an agent with perfect localization and heading estimation (e.g. an agent with an idealized GPS+Compass). However, high-precision localization in indoor environments can not be assumed in realistic settings -- GPS has low precision indoors, (visual) odometry may be noisy, SLAM-based localization can fail, etc. Hence, we will investiage only providing to the agent a fixed relative coordinate for the goal position from the start location. \\[-7pt]
  \item Likewise, the 2019 \habitatchal modeled agent actions (e.g.~\texttt{forward}, \texttt{turn 10$^\circ$ left},...) deterministically. However in real settings, agent intention (e.g.~go forward $1\text{m}$) and the result rarely match perfectly -- actuation error, differing surface materials, and a myriad of other sources of error introduce significant drift over a long trajectory. To model this, we introduce a noise model acquired by benchmarking a real robotic platform~\cite{pyrobot2019}. Visual sensing is an excellent means of combating this ``dead-reckoning'' drift and this change allows participants to study methodologies that are robust to and can correct for this noise.\\[-7pt]
  \item Finally, we will introduce realistic models of sensor noise for RGB and depth sensors -- narrowing the gap between perceptual experiences agents would have in simulation and reality.
\end{compactitem}

We look forward to supporting the community in establishing a benchmark to evaluate the state-of-the-art in methods for embodied navigation agents.

\section{Future Work}

We described the design and implementation of the Habitat platform. 
Our goal is to unify existing community efforts and to accelerate research into embodied AI.
This is a long-term effort that will succeed only by full engagement of the broader research community.

Experiments enabled by the generic dataset support and the high performance of the Habitat stack indicate that i)~learning-based agents can match and exceed the performance of classic visual navigation methods when trained for long enough and ii)~learned agents equipped with depth sensors generalize well between different 3D environment datasets in comparison to agents equipped with only RGB.

\xhdr{Feature roadmap.} 
Our near-term development roadmap will focus on incorporating physics simulation and enabling physics-based interaction between mobile agents and objects in 3D environments.
\habitatsim's scene graph representation is well-suited for integration with physics engines, allowing us to directly control the state of individual objects and agents within a scene graph. 
%
%
Another planned avenue of future work involves procedural generation of 3D environments by leveraging a combination of 3D reconstruction and virtual object datasets.
By combining high-quality reconstructions of large indoor spaces with separately reconstructed or modelled objects, we can take full advantage of our hierarchical scene graph representation to introduce controlled variation in the simulated 3D environments.

Lastly, we plan to focus on distributed simulation settings that involve large numbers of agents potentially interacting with one another in competitive or collaborative scenarios.

\vspace{1em}
\xhdr{Acknowledgments.}
We thank the reviewers for their helpful suggestions.
The Habitat project would not have been possible without the support and contributions of many individuals.
We are grateful to Mandeep Baines, Angel Xuan Chang, Alexander Clegg, Devendra Singh Chaplot, Xinlei Chen, Wojciech Galuba, Georgia Gkioxari, Daniel Gordon, Leonidas Guibas, Saurabh Gupta, Jerry (Zhi-Yang) He, Rishabh Jain, Or Litany, Joel Marcey, Dmytro Mishkin, Marcus Rohrbach, Amanpreet Singh, Yuandong Tian, Yuxin Wu, Fei Xia, Deshraj Yadav, Amir Zamir, and Jiazhi Zhang for their help.

\vspace{1em}
\xhdr{Licenses for referenced datasets.}
\\
{
Gibson: \url{https://storage.googleapis.com/gibson_material/Agreement%20GDS%2006-04-18.pdf}
\\
Matterport3D: \url{http://kaldir.vc.in.tum.de/matterport/MP_TOS.pdf}.
}

\newpage
\balance

{\small
\bibliographystyle{ieee_fullname}
\bibliography{biblio}
}

\newpage
\appendix
\section{Habitat Platform Details}

As described in the main paper, Habitat consists of the following components:
\begin{compactitem}
  \item \habitatsim: a flexible, high-performance 3D simulator with configurable agents, multiple sensors, and generic 3D dataset handling (with built-in support for Matterport3D~\cite{Chang2017}, Gibson~\cite{Xia2018}, and other datasets). \habitatsim is fast -- when rendering a realistic scanned scene from the Matterport3D dataset, \habitatsim achieves several thousand frames per second (fps) running single-threaded, and can reach over $10@000$ fps multi-process on a single GPU.

  \item \habitatapi: a modular high-level library for end-to-end development of embodied AI -- defining embodied AI tasks (\eg navigation~\cite{Anderson2018-Evaluation}, instruction following~\cite{Anderson2018-Language}, question answering~\cite{embodiedqa}), 
  configuring embodied agents (physical form, sensors, capabilities), training these agents (via imitation or reinforcement learning, or via classic SLAM), and benchmarking their performance on the defined tasks using standard metrics~\cite{Anderson2018-Evaluation}. 
  \habitatapi currently uses \habitatsim as the core simulator, 
  but is designed with a modular abstraction for the simulator 
  backend to maintain compatibility over multiple simulators.
\end{compactitem}

\xhdr{Key abstractions.}
The Habitat platform relies on a number of key abstractions that model the domain of embodied agents and tasks that can be carried out in three-dimensional indoor environments.
Here we provide a brief summary of key abstractions:
\begin{compactitem}
    \item \texttt{Agent}: a physically embodied agent with a suite of \texttt{Sensors}. Can observe the environment and is capable of taking actions that change agent or environment state.
    \item \texttt{Sensor}: associated with a specific \texttt{Agent}, capable of returning observation data from the environment at a specified frequency.
    \item \texttt{SceneGraph}: a hierarchical representation of a 3D environment that organizes the environment into regions and objects which can be programmatically manipulated.
    \item \texttt{Simulator}: an instance of a simulator backend. Given actions for a set of configured \texttt{Agents} and \texttt{SceneGraphs}, can update the state of the \texttt{Agents} and \texttt{SceneGraphs}, and provide observations for all active \texttt{Sensors} possessed by the \texttt{Agents}.
\end{compactitem}

These abstractions connect the different layers of the platform. They also enable generic and portable specification of embodied AI tasks.

\begin{figure*}
  \centering
  \includegraphics[width=0.9\textwidth]{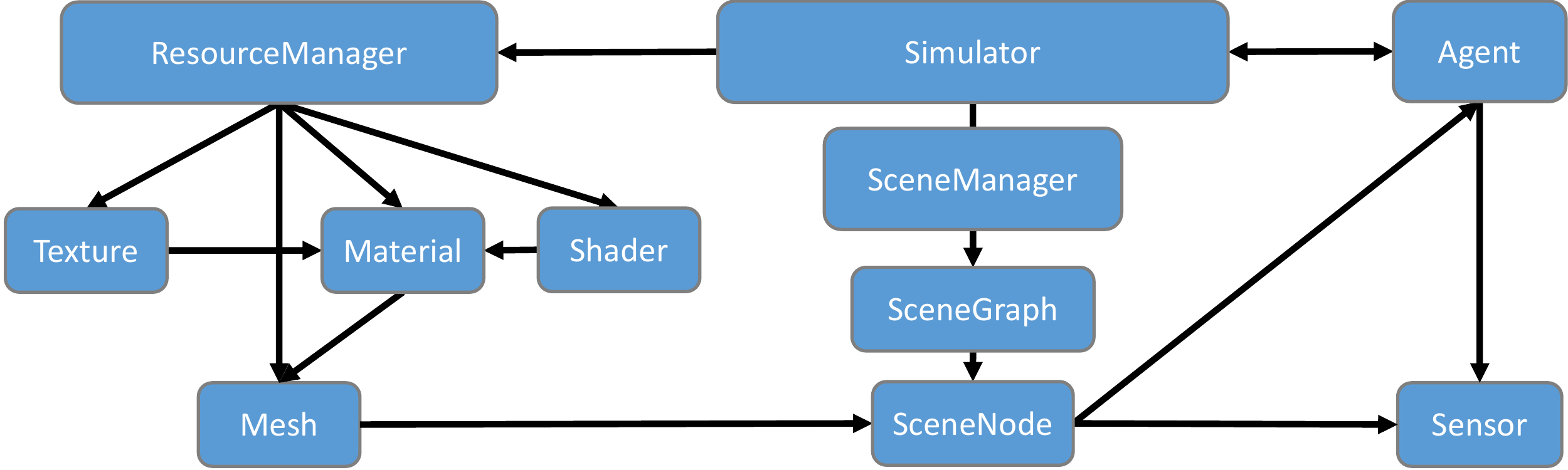}
  \caption{Architecture of \habitatsim main classes. The Simulator delegates management of all resources related to 3D environments to a ResourceManager that is responsible for loading and caching 3D environment data from a variety of on-disk formats. These resources are used within SceneGraphs at the level of individual SceneNodes that represent distinct objects or regions in a particular Scene. Agents and their Sensors are instantiated by being attached to SceneNodes in a particular SceneGraph.}
\label{fig:sim_architecture}
\end{figure*}

\begin{figure}
  \centering
  \includegraphics[width=\linewidth]{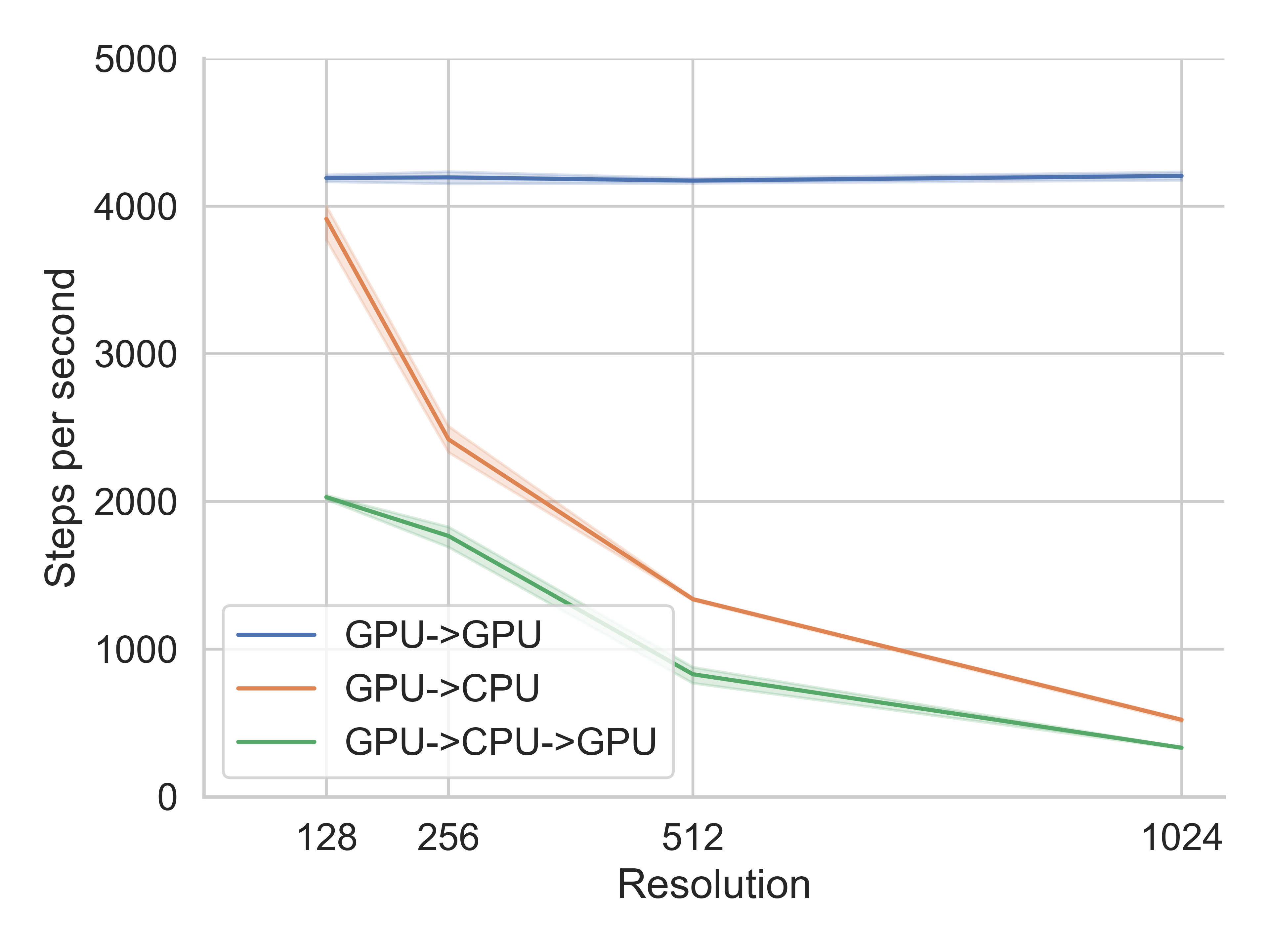}
  \caption{Performance of \habitatsim under different sensor frame memory transfer strategies for increasing image resolution. We see that `GPU->GPU' is unaffected by image resolution while other strategies degrade rapidly.}
\label{fig:benchmark_plots}
\end{figure}

\xhdr{Habitat-Sim.}
The architecture of the \habitatsim backend module is illustrated in \Cref{fig:sim_architecture}.
The design of this module ensures a few key properties:
\begin{compactitem}
  \item Memory-efficient management of 3D environment resources (triangle mesh geometry, textures, shaders) ensuring shared resources are cached and reused.
  \item Flexible, structured representation of 3D environments using \texttt{SceneGraphs}, allowing for programmatic manipulation of object state, and combination of objects from different environments.
  \item High-efficiency rendering engine with multi-attachment render pass to reduce overhead for multiple sensors.
  \item Arbitrary numbers of \texttt{Agents} and corresponding \texttt{Sensors} that can be linked to a 3D environment by attachment to a \texttt{SceneGraph}.
\end{compactitem}
The performance of the simulation backend surpasses that of prior work operating on realistic reconstruction datasets by a large margin. \Cref{tab:performance} reports performance statistics on a test scene from the Matterport3D dataset.
Single-thread performance reaches several thousand frames per second (fps), 
while multi-process operation with several simulation backends can reach over $10@000$ fps on a single GPU.
In addition, by employing OpenGL-CUDA interoperation we enable direct sharing of rendered image frames with ML frameworks such as PyTorch without a measurable impact on performance as the image resolution is increased (see \Cref{fig:benchmark_plots}).

\begin{table*}
  \centering
  \begin{tabular}{@{}lrrrrrrrrr@{}}
      \toprule
      & \multicolumn{3}{c}{GPU$\rightarrow$CPU$\rightarrow$GPU} & \multicolumn{3}{c}{GPU$\rightarrow$CPU} & \multicolumn{3}{c}{GPU$\rightarrow$GPU}\\
      \cmidrule(l){2-4} \cmidrule(l){5-7} \cmidrule(l){8-10}
      Sensors / number of processes & $1$ & $3$ & $5$ & $1$ & $3$ & $5$ & $1$ & $3$ & $5$ \\
      \midrule
      RGB & $2@346$ & $6@049$ & $7@784$ & $3@919$ & $8@810$ & $11@598$ & $4@538$ & $8@573$ & $7@279$\\
      RGB + depth & $1@260$ & $3@025$ & $3@730$ & $1@777$ & $4@307$ & $5@522$ & $2@151$ & $3@557$ & $3@486$\\
      RGB + depth + semantics\footnotemark{} 
      & $378$ & $463$ & $470$ & $396$ & $465$ & $466$ & $464$ & $455$ & $453$\\
      \bottomrule
  \end{tabular}
  \vspace{1mm}
  \caption{Performance of \habitatsim in frames per second for an example Matterport3D scene (id 17DRP5sb8fy) on a Xeon E5-2690 v4 CPU and Nvidia Titan Xp GPU, measured at a frame resolution of 128x128, under different frame memory transfer strategies and with a varying number of concurrent simulator processes sharing the GPU.
  `GPU-CPU-GPU' indicates passing of rendered frames from OpenGL context to CPU host memory and back to GPU device memory for use in optimization, `GPU-CPU' only reports copying from OpenGL context to CPU host memory, whereas `GPU-GPU' indicates direct sharing through OpenGL-CUDA interoperation.}
  \label{tab:performance}
\end{table*}


\footnotetext{Note: The semantic sensor in Matterport3D requires using additional 3D meshes with significantly more geometric complexity, leading to reduced performance. We expect this to be addressed in future versions, leading to speeds comparable to RGB + depth.}

\xhdr{Habitat-API.}
The second layer of the Habitat platform (\habitatapi) focuses on creating a general and flexible API for defining embodied agents, tasks that they may carry out, and evaluation metrics for those tasks.
When designing such an API, a key consideration is to allow for easy extensibility of the defined abstractions.
This is particularly important since many of the parameters of embodied agent tasks, specific agent configurations, and 3D environment setups can be varied in interesting ways.
Future research is likely to propose new tasks, new agent configurations, and new 3D environments.

\begin{figure*}[t]
  \centering
  \includegraphics[width=\textwidth]{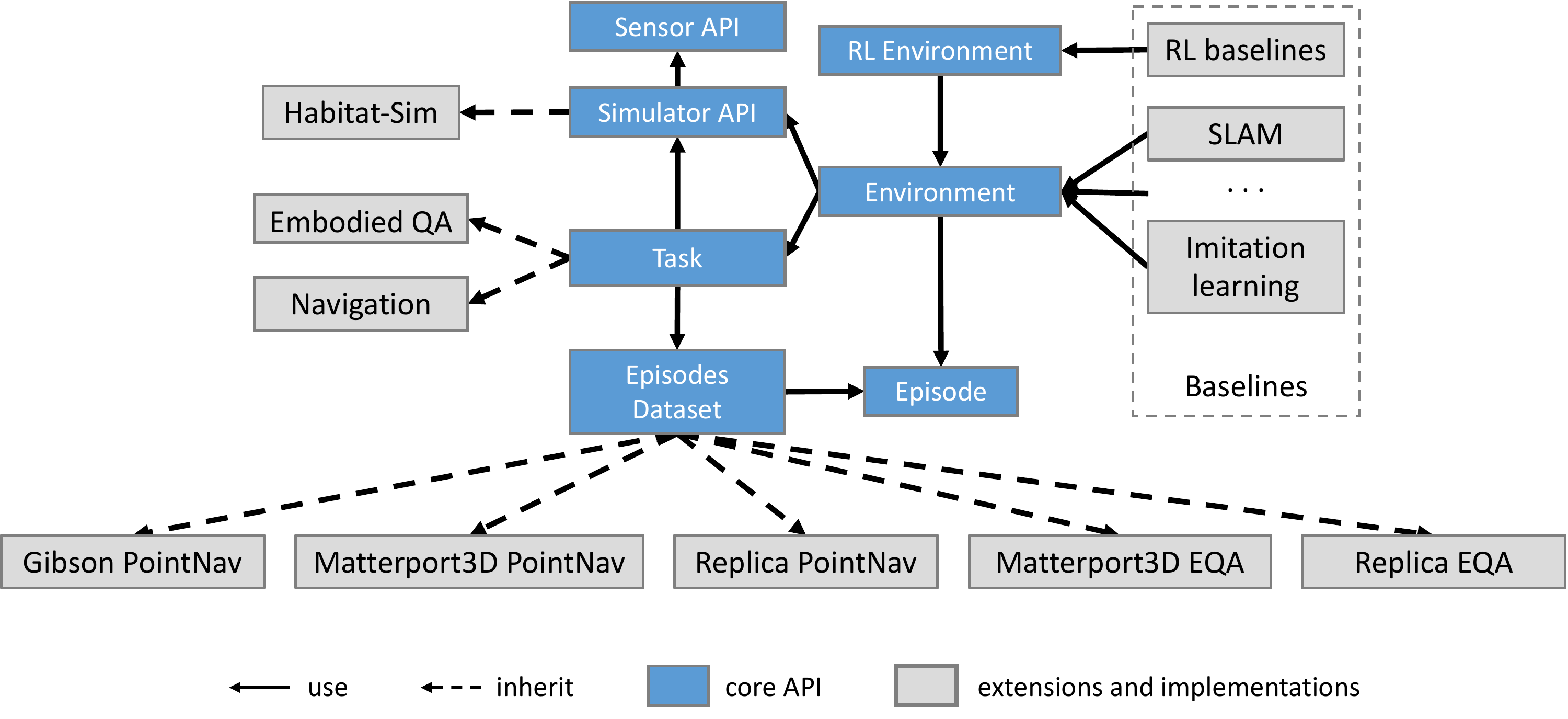}
  \caption{Architecture of \habitatapi. The core functionality defines fundamental building blocks such as the API for interacting with the simulator backend and receiving observations through \texttt{Sensors}. Concrete simulation backends, 3D datasets, and embodied agent baselines are implemented as extensions to the core API.}
\label{fig:api_architecture}
\end{figure*}

The API allows for alternative simulator backends to be used, beyond the \habitatsim module that we implemented.
This modularity has the advantage of allowing incorporation of existing simulator backends to aid in transitioning from experiments that previous work has performed using legacy frameworks.
The architecture of \habitatapi is illustrated in \Cref{fig:api_architecture}, indicating core API functionality and functionality implemented as extensions to the core.

Above the API level, we define a concrete embodied task such as visual navigation.
This involves defining a specific dataset configuration, specifying the structure of episodes (\eg number of steps taken, termination conditions), training curriculum (progression of episodes, difficulty ramp), and evaluation procedure (\eg test episode sets and task metrics).
An example of loading a pre-configured task (PointNav)
and stepping through the environment with a random agent 
is shown in the code below.

\section{Additional Dataset Statistics}

\begin{figure*}[t]
  \centering
  \includegraphics[width=\textwidth]{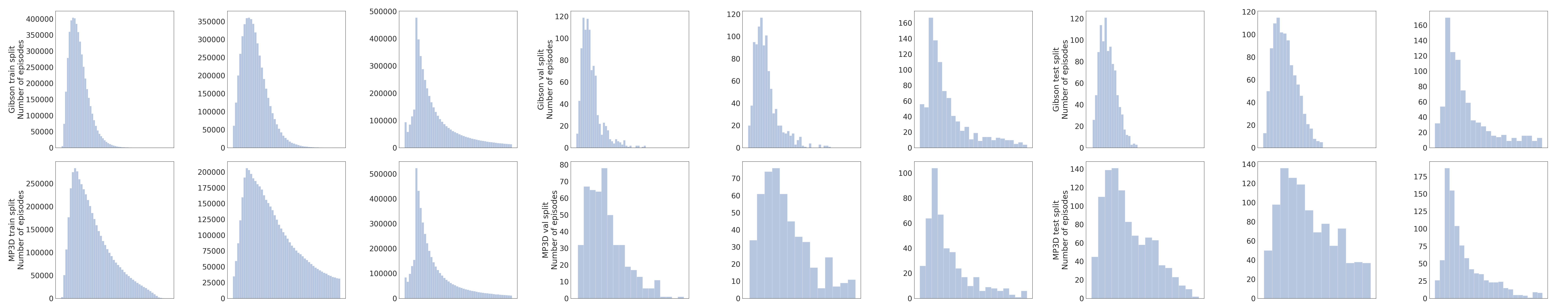}
  \caption{Statistics of PointGoal navigation episodes. From left: distribution over Euclidean distance between start and goal, distribution over geodesic distance along shortest path between start and goal, and distribution over the ratio of geodesic to Euclidean distance.
  }
\label{fig:dataset_stats}
\end{figure*}

In \Cref{tab:dataset_stats} we summarize the train, validation and test split sizes for all three datasets used in our experiments.
We also report the average geodesic distance along the shortest path (\gdsp) between starting point and goal position.
As noted in the main paper, Gibson episodes are significantly shorter than Matterport3D ones.
\Cref{fig:dataset_stats} visualizes the episode distributions over geodesic distance (\gdsp), Euclidean distance between start and goal position, and the ratio of the two (an approximate measure of complexity for the episode).
We again note that Gibson episodes have more episodes with shorter distances, leading to the dataset being overall easier than the Matterport3D dataset.

\inputpython{api-example_single_col.py}{1}{17}

\begin{table}
\centering
\footnotesize
\begin{tabular}{@{}lccc@{}}
  \toprule
  Dataset & scenes (\#) & episodes (\#) & average \gdsp (\text{m})\\
  \midrule
  Matterport3D & 58 / 11 / 18 & 4.8M / 495 / 1008 & 11.5 / 11.1 / 13.2 \\
  Gibson & 72 / 16 / 10        & 4.9M / 1000 / 1000   & 6.9 / 6.5 / 7.0 \\
  \bottomrule
\end{tabular}\\[3pt]
\caption{Statistics of the PointGoal navigation datasets that we precompute for the Matterport3D and Gibson datasets: total number of scenes, total number of episodes, and average geodesic distance between start and goal positions. Each cell reports train / val / test split statistics.}
\label{tab:dataset_stats}
\end{table}

\begin{table}
\centering
\footnotesize
\begin{tabular}{@{}lcccc@{}}
  \toprule
  Dataset & Min & Median & Mean & Max \\
  \midrule
  Matterport3D & 18 & 90.0 & 97.1 & 281 \\
  Gibson & 15 & 60.0 & 63.3 & 207\\
  \bottomrule
\end{tabular}
\caption{Statistics of path length (in actions) for an oracle which greedily fits actions to follow the negative of geodesic distance gradient on the PointGoal navigation validation sets.  This provides expected horizon lengths for a near-perfect agent and contextualizes the decision for a max-step limit of 500.}
\label{tab:dataset_stats}
\end{table}

\section{Additional Experimental Results}

In order to confirm that the trends we observe for the experimental results presented in the paper hold for much larger amounts of experience, we scaled our experiments to 800M steps.
We found that (1) the ordering of the visual inputs stays \depth $>$ \rgbd $>$ \rgb $>$ \blind; (2) \rgb is consistently better than \blind (by $0.06$/$0.03$ \spl on Gibson/Matterport3D), and (3) \rgbd outperforms SLAM on Matterport3D (by $0.16$ \spl). 

\begin{figure}
  \centering
  \includegraphics[width=\columnwidth]{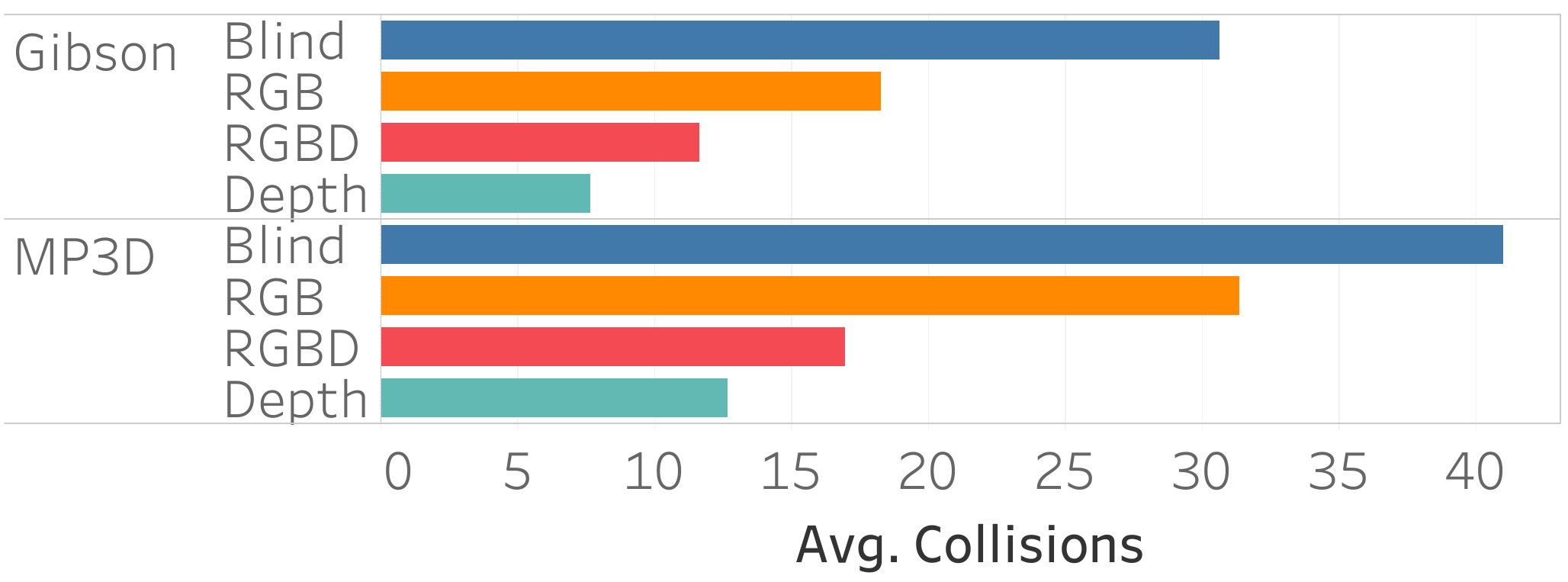}
  \caption{Average number of collisions during successful navigation episodes for the different sensory configurations of the RL (PPO) baseline agent on test set episodes for the Gibson and Matterport3D datasets. The \blind agent experiences the highest number of collisions, while agents possessing depth sensors (\depth and \rgbd) have the fewest collisions on average.}
  \label{fig:collisions}
  \vspace{\captionReduceBot}
\end{figure}

\subsection{Analysis of Collisions}

To further characterize the behavior of learned agents during navigation we plot the average number of collisions in \Cref{fig:collisions}.
We see that \blind incurs a much larger number of collisions than other agents, providing evidence for `wall-following' behavior.
Depth-equipped agents have the lowest number of collisions, while \rgb agents are in between.

\subsection{Noisy Depth}

To investigate the impact of noisy depth measurements on agent performance, we re-evaluated depth agents (without re-training) on noisy depth generated using a simple noise model: iid Gaussian noise ($\mu=0$, $\sigma=0.4$) at each pixel in inverse depth (larger depth = more noise).
We observe a drop of $0.13$ and $0.02$ SPL for depth-RL and SLAM on Gibson-val (depth-RL still outperforms SLAM).
Note that SLAM from ~\cite{mishkin2019benchmarking} utilizes ORB-SLAM2, which is quite robust to noise, while depth-RL was trained without noise.
If we increase $\sigma$ to $0.1$, depth-RL gets $0.12$ SPL whereas SLAM suffers catastrophic failures.

\section{Gibson Dataset Curation}

\begin{figure*}
    \centering
    \captionsetup[subfigure]{justification=centering,labelformat=empty}
    \begin{subfigure}[b]{0.49\textwidth}
      \includegraphics[width=\textwidth]{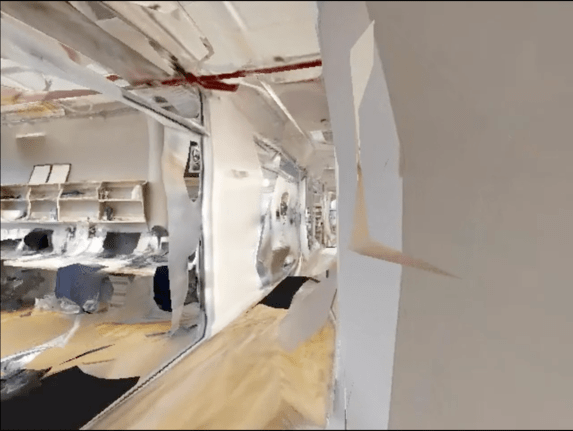}
      \caption{$0$: critical reconstruction artifacts, holes, or texture issues}
    \end{subfigure}
    \begin{subfigure}[b]{0.49\textwidth}
        \includegraphics[width=\textwidth]{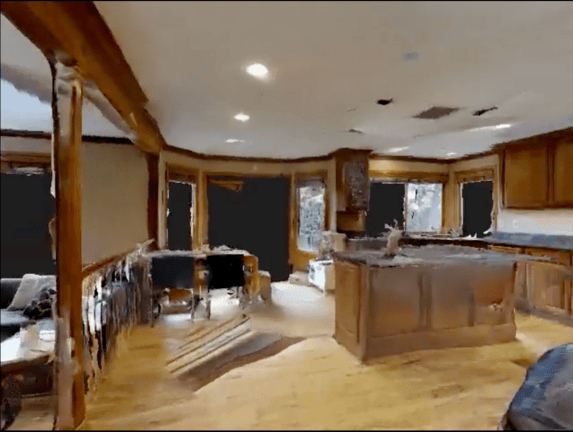}
        \caption{$1$: big holes or significant texture issues and reconstruction artifacts}
    \end{subfigure}
    \begin{subfigure}[b]{0.49\textwidth}
        \includegraphics[width=\textwidth]{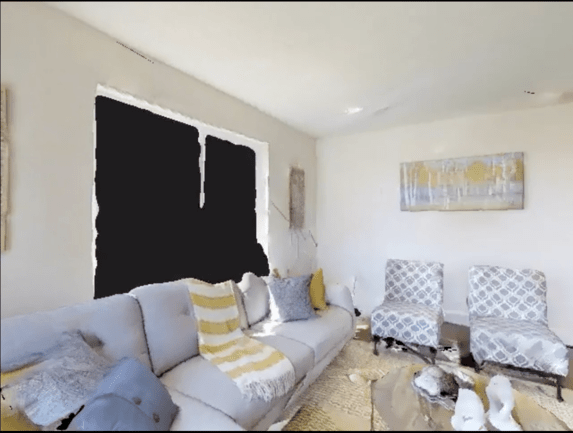}
        \caption{$2$: big holes or significant texture issues, but good reconstruction}
    \end{subfigure}
    \begin{subfigure}[b]{0.49\textwidth}
        \includegraphics[width=\textwidth]{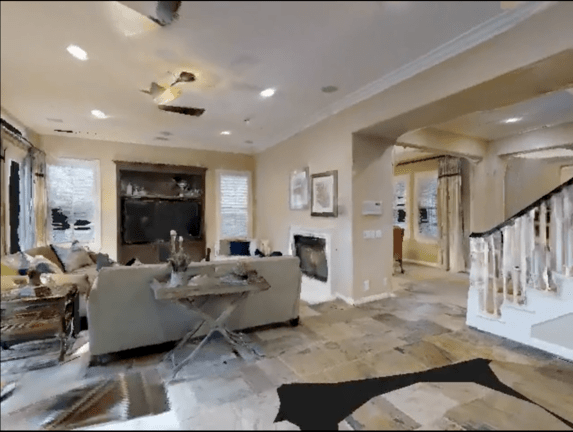}
        \caption{$3$: small holes, some texture issues, good reconstruction}
    \end{subfigure}
    \begin{subfigure}[b]{0.49\textwidth}
        \includegraphics[width=\textwidth]{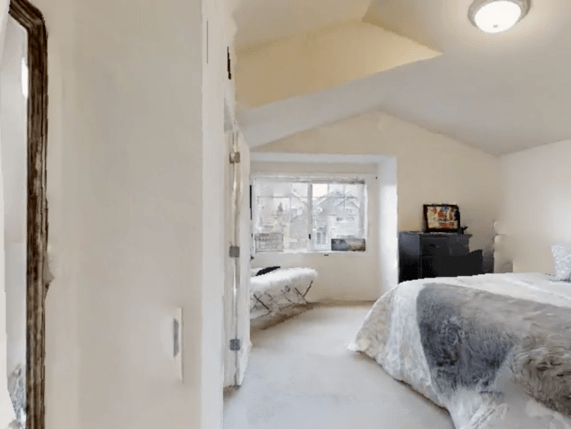}
        \caption{$4$: no holes, some texture issues, good reconstruction}
    \end{subfigure}
    \begin{subfigure}[b]{0.49\textwidth}
        \includegraphics[width=\textwidth]{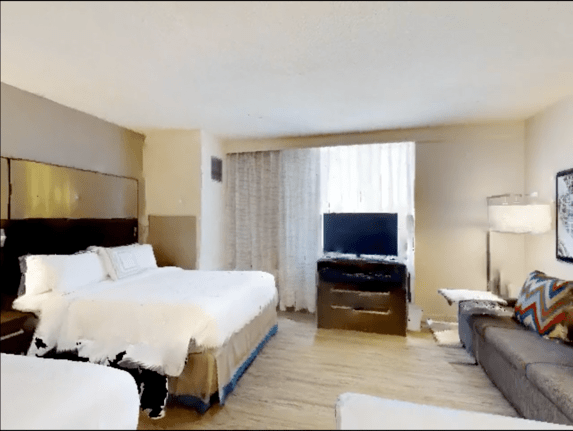}
        \caption{$5$: no holes, uniform textures, good reconstruction}
    \end{subfigure}
    \caption{Rating scale used in curation of 3D textured mesh reconstructions from the Gibson dataset. We use only meshes with ratings of $4$ or higher for the Habitat Challenge dataset.}
    \label{fig:gibson-ratings}
\end{figure*}

We manually curated the full dataset of Gibson 3D textured meshes~\cite{Xia2018} to select meshes that do not exhibit significant reconstruction artifacts such as holes or texture quality issues.
A key issue that we tried to avoid is the presence of holes or cracks in floor surfaces.
This is particularly problematic for navigation tasks as it divides seemingly connected navigable areas into non-traversable disconnected components.
We manually annotated the scenes (using the $0$ to $5$ quality scale shown in \Cref{fig:gibson-ratings}) and only use scenes with a rating of $4$ or higher, i.e., no holes, good reconstruction, and negligible texture issues to generate the dataset episodes.

\section{Reproducing Experimental Results}
Our experimental results can be reproduced using the \habitatapi (commit \href{https://github.com/facebookresearch/habitat-api/tree/ec9557a3623991208a80f836fe557f8028209297}{ec9557a}) and \habitatsim (commit \href{https://github.com/facebookresearch/habitat-sim/tree/d383c2011bf1baab2ce7b3cd40aea573ad2ddf71}{d383c20}) repositories. The code for running experiments is present under the folder \texttt{habitat-api/habitat\_baselines}. Below is the shell script we used for our RL experiments:

\inputpython{ppo_train.sh}{1}{17}

For running SLAM please refer to \href{https://github.com/facebookresearch/habitat-api/tree/ec9557a3623991208a80f836fe557f8028209297/habitat_baselines/slambased}{habitat-api/habitat\_baselines/slambased}.


\section{Example Navigation Episodes}

\Cref{fig:nav_episodes} visualizes additional example navigation episodes for the different sensory configurations of the RL (PPO) agents that we describe in the main paper.
\blind agents have the lowest performance, colliding much more frequently with the environment and adopting a `wall hugging' strategy for navigation.
\rgb agents are less prone to collisions but still struggle to navigate to the goal position successfully in some cases.
In contrast, depth-equipped agents are much more efficient, exhibiting fewer collisions, and navigating to goals more successfully (as indicated by the overall higher \spl values).

\begin{figure*}
  \centering
  \captionsetup[subfigure]{justification=centering,labelformat=empty}
  \begin{subfigure}[b]{\textwidth}
    \captionsetup[subfigure]{justification=centering,labelformat=empty}
    \caption{\large Gibson}
    \begin{subfigure}[b]{0.49\textwidth}
    \includegraphics[width=\textwidth]{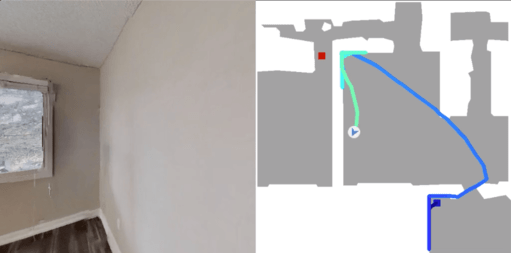}
    \caption{\blind \spl $=0.00$}
    \end{subfigure}
    \begin{subfigure}[b]{0.49\textwidth}
      \includegraphics[width=\textwidth]{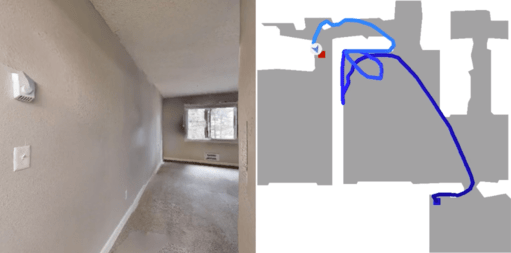}
      \caption{\rgb \spl $=0.45$}
    \end{subfigure}
    \begin{subfigure}[b]{0.49\textwidth}
      \includegraphics[width=\textwidth]{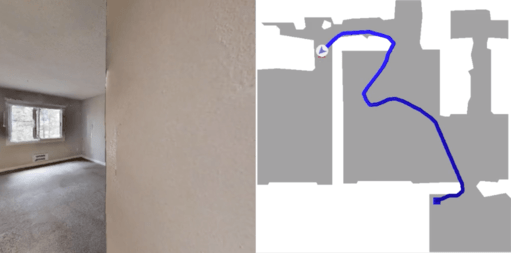}
      \caption{\rgbd \spl $=0.82$}
    \end{subfigure}
    \begin{subfigure}[b]{0.49\textwidth}
      \includegraphics[width=\textwidth]{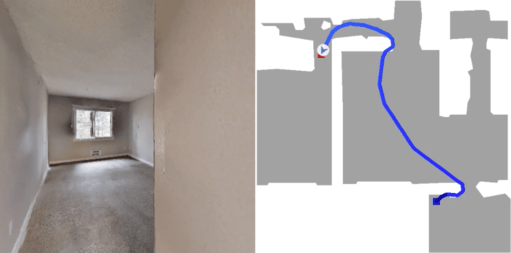}
      \caption{\depth \spl $=0.88$}
    \end{subfigure}
    \begin{subfigure}[b]{0.49\textwidth}
      \includegraphics[width=\textwidth]{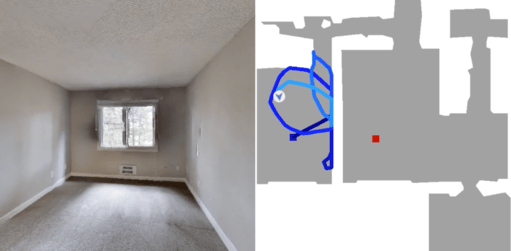}
      \caption{\blind \spl $=0.00$}
      \end{subfigure}
      \begin{subfigure}[b]{0.49\textwidth}
        \includegraphics[width=\textwidth]{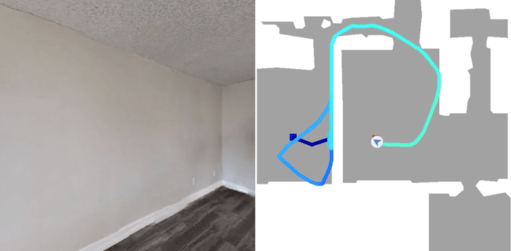}
        \caption{\rgb \spl $=0.29$}
      \end{subfigure}
      \begin{subfigure}[b]{0.49\textwidth}
        \includegraphics[width=\textwidth]{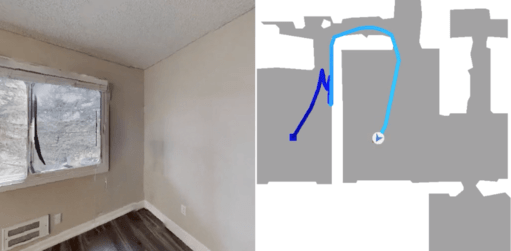}
        \caption{\rgbd \spl $=0.49$}
      \end{subfigure}
      \begin{subfigure}[b]{0.49\textwidth}
        \includegraphics[width=\textwidth]{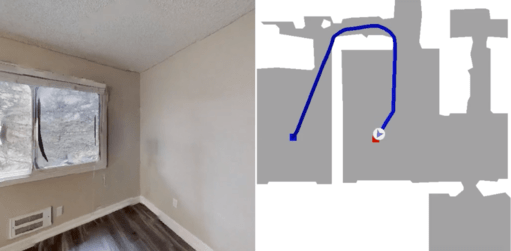}
        \caption{\depth \spl $=0.96$}
      \end{subfigure}
  \end{subfigure}
  \caption{Additional navigation example episodes for the different sensory configurations of the RL (PPO) agent, visualizing trials from the Gibson and MP3D val sets. A \textbf{blue dot} and \textbf{red dot} indicate the starting and goal positions, and the \textbf{blue arrow} indicates final agent position. The \textbf{blue-green-red line} is the agent's trajectory. Color shifts from blue to red as the maximum number of allowed agent steps is approached.}
  \label{fig:nav_episodes}
\end{figure*}

\begin{figure*}
  \ContinuedFloat
  \centering
  \captionsetup[subfigure]{justification=centering,labelformat=empty}
  \begin{subfigure}[b]{\textwidth}
    \captionsetup[subfigure]{justification=centering,labelformat=empty}
    \caption{\large MP3D}
    \begin{subfigure}[b]{0.49\textwidth}
      \includegraphics[width=\textwidth]{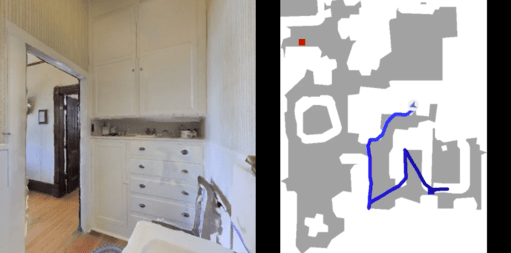}
      \caption{\blind \spl $=0.00$}
    \end{subfigure}
    \begin{subfigure}[b]{0.49\textwidth}
      \includegraphics[width=\textwidth]{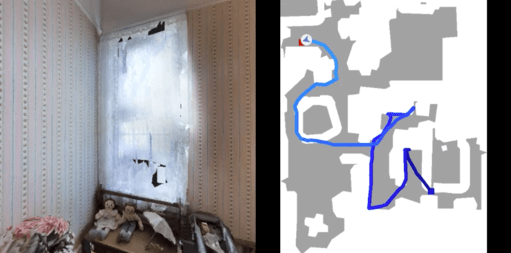}
      \caption{\rgb \spl $=0.40$}
    \end{subfigure}
    \begin{subfigure}[b]{0.49\textwidth}
      \includegraphics[width=\textwidth]{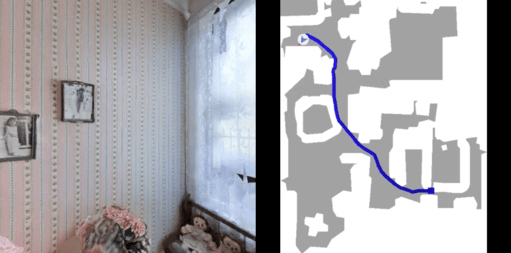}
      \caption{\rgbd \spl $=0.92$}
    \end{subfigure}
    \begin{subfigure}[b]{0.49\textwidth}
      \includegraphics[width=\textwidth]{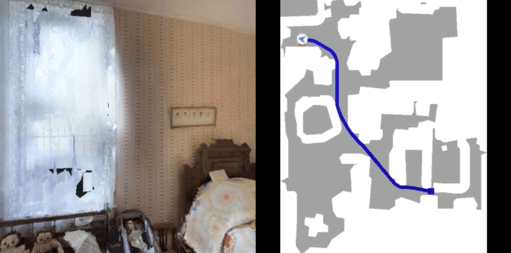}
      \caption{\depth \spl $=0.98$}
    \end{subfigure}
  \end{subfigure}
  \caption{Additional navigation example episodes for the different sensory configurations of the RL (PPO) agent, visualizing trials from the Gibson and MP3D val sets. A \textbf{blue dot} and \textbf{red dot} indicate the starting and goal positions, and the \textbf{blue arrow} indicates final agent position. The \textbf{blue-green-red line} is the agent's trajectory. Color shifts from blue to red as the maximum number of allowed agent steps is approached.}
\end{figure*}

\end{document}